\documentclass[final]{cvpr}

\usepackage{times}
\usepackage{epsfig}
\usepackage{graphicx}
\usepackage{amsmath}
\usepackage{amssymb}
\usepackage[utf8]{inputenc}
\usepackage[T1]{fontenc}
\usepackage{url}
\usepackage{booktabs}
\usepackage{amsfonts}
\usepackage{nicefrac}
\usepackage{microtype}
\usepackage{times,helvet,courier}
\usepackage[tight,footnotesize]{subfigure}
\usepackage{amsthm,bm}
\usepackage{multirow,makecell,footmisc}
\usepackage{color}
\usepackage{bbm}
\usepackage{pifont}

\usepackage{algorithm}
\usepackage[noend]{algpseudocode}
\usepackage{setspace}

\usepackage{marvosym}

\newcommand{\ymark}{\ding{51}}%
\newcommand{\xmark}{\ding{55}}%

\newcommand{\myparagraph}[1]{\vspace{-5pt}\paragraph{#1}}

\usepackage[pagebackref=true,breaklinks=true,colorlinks,bookmarks=false]{hyperref}

\def\cvprPaperID{5777} 
\def\confYear{CVPR 2021}

\begin{document}


\title{Open Domain Generalization with Domain-Augmented Meta-Learning}

\author{Yang Shu\thanks{Equal contribution.} , Zhangjie Cao\footnotemark[1] , Chenyu Wang, Jianmin Wang, Mingsheng Long (\Letter)\\
School of Software, BNRist, Tsinghua University, China\\
{\tt\small \{shu-y18,caozj14,cy-wang18\}@mails.tsinghua.edu.cn, \{jimwang,mingsheng\}@tsinghua.edu.cn}
}

\maketitle

\begin{abstract}
Leveraging datasets available to learn a model with high generalization ability to unseen domains is important for computer vision, especially when the unseen domain's annotated data are unavailable.  We study a novel and practical problem of Open Domain Generalization (OpenDG), which learns from different source domains to achieve high performance on an unknown target domain, where the distributions and label sets of each individual source domain and the target domain can be different. The problem can be generally applied to diverse source domains and widely applicable to real-world applications. We propose a Domain-Augmented Meta-Learning framework to learn open-domain generalizable representations. We augment domains on both feature-level by a new Dirichlet mixup and label-level by distilled soft-labeling, which complements each domain with missing classes and other domain knowledge. We conduct meta-learning over domains by designing new meta-learning tasks and losses to preserve domain unique knowledge and generalize knowledge across domains simultaneously. Experiment results on various multi-domain datasets demonstrate that the proposed Domain-Augmented Meta-Learning (DAML) outperforms prior methods for unseen domain recognition.
\end{abstract}

\vspace{-10pt}

\section{Introduction}
\begin{figure}[htbp]
  \centering
  \includegraphics[width=0.45\textwidth]{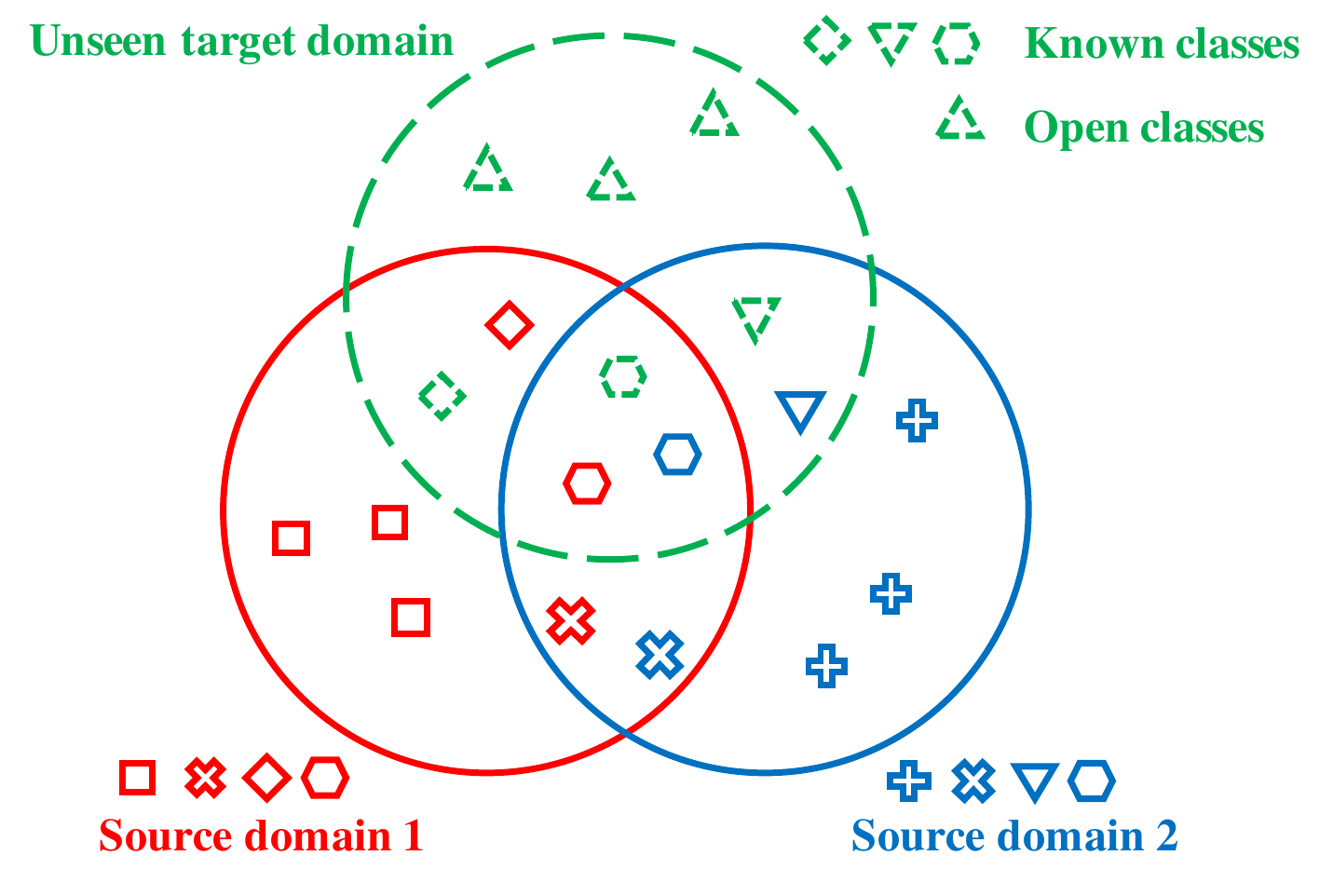}
  \vspace{-5pt}
  \caption{Open Domain Generalization (OpenDG). Different source domains hold disparate label sets. The goal is to learn generalizable representations from these source domains to help classify the known classes and detect open classes in the unseen target domain.}
   \label{fig:setting}
\end{figure}

Deep convolutional neural networks have achieved state-of-the-art performance on wide ranges of computer vision applications with access to large-scale labeled data~\cite{cite:NIPS12CNN,cite:CVPR16DRL,cite:NIPS15FasterRCNN,cite:ICCV17MaskRCNN}. However, for a target domain of interest, collecting enough training data is prohibitive. A practical solution is to generalize the model learned on the existing data to the unseen domain. Since the existing source datasets for training may be from different resources, they may fall into different domains and hold different label sets, e.g., ImageNet~\cite{deng2009imagenet} and DomainNet~\cite{peng2019moment}. Besides, the target domain is totally unknown, and may also have a distribution shift and a different label set from the source domains. We call the valuable and challenging problem as \textbf{Open Domain Generalization (OpenDG)}, where we need to learn generalizable representation from disparate source domains that generalizes well to any unseen target domain, as illustrated in Figure~\ref{fig:setting}. 

There are two key challenges for open domain generalization. (1) Distinct source domains and the unseen target domain are drawn from different distributions with a large distribution shift. (2) The different label sets of distinct source domains cause some classes to exist in many more domains than other classes. The data of minor classes existing in few domains are lacking in diversity. This makes the problem extremely difficult for existing methods \cite{cite:AAAI18MLDG,cite:ECCV18CIAN}. 

\begin{table*}[ht]
    \centering
    \caption{Comparison of the proposed generalization setting with the previous settings related to cross-domain learning. The columns list assumptions made by the problem settings. \textbf{Note that more ``\xmark'' means the method needs less assumption and thus is more widely-applicable.} We can observe that the proposed open domain generalization problem requires no assumptions on the label set, no target data, and no post-training on target data, which is the most general problem setting. \textbf{S} means source while \textbf{T} means target. Note that ``Same between \textbf{S}\&\textbf{T} Domains'' means the union of all source domain label sets equals the target label set, i.e., whether there are open classes.}\label{tab:related_work}
    \resizebox{1.\textwidth}{!}{
    \begin{tabular}{l|cc|cc|c|}
    \toprule
       \multirow{2}{80pt}{Problem Setting} & \multicolumn{2}{|c}{Label Set} & \multicolumn{2}{|c|}{Target Data for Training} & \multirow{2}{80pt}{Post-Training on Target Labeled Data}\\
        \cmidrule{2-3}\cmidrule{4-5}
         & Same for \textbf{S} Domains & Same between \textbf{S}\&\textbf{T} Domains & Labeled Data & Unlabeled Data & \\ 
         \midrule
         Domain Adaptation~\cite{cite:ICML15DAN,long2018conditional} & \ymark & \ymark & \xmark & \ymark & \xmark\\
         Domain Adaptation with Category Shift~\cite{cite:ICCV17OpenSet,cite:CVPR18SAN,you2019universal} & \ymark & \xmark & \xmark & \ymark & \xmark\\
         Multi-Source Domain Adaptation~\cite{zhao2018adversarial} & \ymark & \ymark & \xmark & \ymark & \xmark\\
         Multi-Source Domain Adaptation with Category Shift~\cite{xu2018deep} & \xmark & \ymark & \xmark & \ymark & \xmark \\
         Domain Generalization~\cite{cite:ICML13DICA} & \ymark & \ymark & \xmark & \xmark & \xmark\\
         Heterogeneous Domain Generalization~\cite{li2019feature} & \xmark & \xmark & \xmark & \xmark & \ymark \\
         The Proposed Open Domain Generalization & \xmark & \xmark & \xmark & \xmark & \xmark\\
         \bottomrule
    \end{tabular}}
    \vspace{-12pt}
\end{table*}

To address the first challenge, previous works minimize the distribution distance between domains by adversarial learning~\cite{cite:ICML13DICA,cite:ECCV18CIAN}, which successfully closes the domain gap when all source domains share the same label set. However, according to the second challenge, the different label sets between domains cause these distribution alignment methods to suffer from severe mismatch of classes. For the second challenge, a straightforward way is to manually sample data of minor classes existing in few domains, but the diversity in domains of the class is still limited. The generalization on the minor class is still inferior to other classes. 

To generalize from \emph{arbitrary} source domains to an unseen target domain, we propose a \textbf{Domain-Augmented Meta-Learning (DAML)} framework. To close the domain gap between disparate source domains, we avoid distribution matching but learn generalizable representations across domains by meta-learning. To overcome the disparate label sets of open domain generalization, we propose two domain augmentation methods at both feature-level and label-level. At feature-level, we design a novel Dirichlet mixup (Dir-mixup) to compensate for the missing labels. At label-level, we utilize the soft-labeling distilled from other domains' networks to transfer the knowledge of other domains to the current network. DAML learns a representation that embeds the knowledge of all source domains and is highly generalizable to the unseen target domain. We use the ensemble of all source domain network outputs as the final prediction, which naturally calibrates the predictive uncertainty. In summary:
\begin{itemize}
    \item We propose a new and practical problem: \textbf{Open Domain Generalization (OpenDG)}, which learns from arbitrary source domains with disparate distributions and label sets to generalize to an unseen target domain.
    \item We propose a principled \textbf{Domain-Augmented Meta-Learning (DAML)} framework to address open domain generalization. We augment each domain with novel Dir-mixup and distilled soft-labeling to overcome the disparate label sets of source domains and conduct meta-learning across augmented domains to learn open-domain generalizable representations.
    \item Experiment results on several multi-domain datasets show that compared to previous generalization methods, DAML achieves higher classification accuracy on both known classes and open classes in an unseen target domain even with extremely diverse source domains.
\end{itemize}

\begin{algorithm*}[t]
\setstretch{1.15}
\caption{Training process of Domain-Augmented Meta-Learning (DAML)}
\label{alg}
\begin{algorithmic}[1]
\Require Source datasets $\mathcal{D}_1,\mathcal{D}_2,\cdots,\mathcal{D}_S$, learning rates $\eta$ and $\beta$, Dir-mixup hyper-parameters $\alpha_{\text{max}}$ and $\alpha_{\text{min}}$

\State \textbf{Initialize} $\ \theta_s|_{s=1}^{S}$
\While{Not Converged}
%
\State Sample a batch of data $\mathcal{B}^{\text{tr}}=\{(\mathbf{x}_1, \mathbf{y}_1),(\mathbf{x}_2, \mathbf{y}_2),\cdots,(\mathbf{x}_S, \mathbf{y}_S)\}$ from all source domains $\mathcal{D}_1$, $\mathcal{D}_2$, $\cdots$, $\mathcal{D}_S$.
\For{$s=1, \ldots, S$} \Comment{Meta-training starts}
\State $\bm{\alpha}_s^{\text{tr}} \leftarrow \left\{\alpha_{\text{max}}, \alpha_{\text{min}}, s\right\}$ \Comment{Dir-mixup parameter for meta-training}
\State $\mathcal{B}_s^{\text{D-mix}}=\{(\mathbf{z}^{\text{D-mix}}_s,\mathbf{y}^{\text{D-mix}}_s)\}\leftarrow \text{Dir-mixup}(\{\bm{\alpha}_s^{\text{tr}},\mathcal{B}^{\text{tr}}\})$ \Comment{Obtain Dir-mixup according to Eqn.~\eqref{eqn:mixup}}
\State $\mathcal{B}_s^{\text{distill}}=\{(\mathbf{x}_s, \mathbf{y}_s^{\text{distill}})\}\leftarrow \{G_j|_{j\neq s},F_j|_{j\neq s},\mathcal{B}^{\text{tr}}\}$ \Comment{Obtain distilled soft-label according to Eqn.~\eqref{eqn:distill}}
\State $\mathcal{L}^{\text{tr}}_{s} \leftarrow$ $\left\{G_s(F_s(\mathbf{x}_s)),\mathbf{y}_s,G_s(\mathbf{z}^{\text{D-mix}}_s),\mathbf{y}^\text{D-mix}_s,\mathbf{y}_s^{\text{distill}}\right\}$ using data in $\mathcal{B}^{\text{tr}}$, $\mathcal{B}^{\text{D-mix}}_s$, and $\mathcal{B}^{\text{distill}}_s$ \Comment{According to
Eqn~\eqref{eqn:meta-train}}
\State $\theta_{F_s^\prime,G_s^\prime} = \theta_{F_s,G_s}-\eta \nabla_\theta \mathcal{L}_{s}^\text{tr}$ 
\EndFor

\State Sample another batch of data $\mathcal{B}^{\text{obj}}=\{(\mathbf{x}_1, \mathbf{y}_1),(\mathbf{x}_2, \mathbf{y}_2),\cdots,(\mathbf{x}_S, \mathbf{y}_S)\}$ from all source domains $\mathcal{D}_1$, $\mathcal{D}_2$, $\cdots$, $\mathcal{D}_S$.
\For{$s=1, \ldots, S$} \Comment{Meta-objective starts}
\State $\bm{\alpha}_s^{\text{obj}} \leftarrow \left\{\alpha_{\text{min}}, \alpha_{\text{max}}, s\right\}$ \Comment{Dir-mixup parameter for meta-objective}
\State ${\mathcal{B}_s^{\text{D-mix}}}^\prime=\{({\mathbf{z}^{\text{D-mix}}_s}^\prime,{\mathbf{y}^{\text{D-mix}}_s}^\prime)\}\leftarrow \text{Dir-mixup}(\{\bm{\alpha}_s^{\text{obj}},\mathcal{B}^{\text{obj}}\})$ \Comment{Obtain Dir-mixup according to Eqn.~\eqref{eqn:mixup}}

\State $\mathcal{L}^{\text{obj}}_{s} \leftarrow$ $\left\{G^\prime_s(F^\prime_s(\mathbf{x}_j))|_{j\neq s},\mathbf{y}_j|_{j\neq s},G^\prime_s({\mathbf{z}^{\text{D-mix}}_s}^\prime),{\mathbf{y}^\text{D-mix}_s}^\prime\right\}$ using data in $\mathcal{B}^{\text{obj}}$ and ${\mathcal{B}_s^{\text{D-mix}}}^\prime$ \Comment{According to
Eqn~\eqref{eqn:meta-val}}
\State $\theta_{F_s,G_s}\leftarrow \theta_{F_s,G_s}-\beta \nabla_\theta (\mathcal{L}^{\text{tr}}_{s}+ \mathcal{L}_{s}^{\text{obj}})$ \Comment{Update parameters with meta-learning}
\EndFor

\EndWhile
\State \textbf{return} $\ \theta_s|_{s=1}^{S}$
\end{algorithmic}
\end{algorithm*}

\section{Related Work}
In this section, we briefly discuss works related to ours, including domain adaptation, domain generalization, and data augmentation methods. We compare our problem setting with the problem settings of previous works in Table~\ref{tab:related_work}.

\textbf{Domain Adaptation} aims to adapt the model from the source domain to the target domain, which typically mitigates the domain gap by minimizing the distribution distance~\cite{cite:ICML15DANN,long2018conditional}. However, the classic domain adaptation requires the same label set between source and target domains. Recent works try to extend domain adaptation to varied source and target label sets~\cite{cite:CVPR18SAN,cite:ICCV17OpenSet,cite:ECCV18OpenSetBack,you2019universal}, but the solution relies on the target unlabeled data, which is not available in the open domain generalization setting. 

Multi-source domain adaptation is more related to our work with more than one source domain. Most of the works assume that all the source domains share the same label set~\cite{zhao2018adversarial,peng2019moment}, which can be easily violated in practice since source domains may be drawn from different resources. DCN~\cite{xu2018deep} moves a step forward to remove the constraint on the source label sets but still requires the union of source label sets to be the same as the target label set. We instead require no label set constraint and no target data for training.

\textbf{Domain Generalization} aims to learn a generalizable model with only source data to achieve high performance in an unseen target domain~\cite{cite:ECCV12undoing,cite:ICML13DICA}, which typically learns domain-invariant features across source domains~\cite{cite:ICML13DICA,cite:ICCV15MTA,cite:TPAMI2017SCA,cite:CVPR18AFL,cite:Arxiv18ADG,cite:ICML20CSD,cite:ECCV20DMG}. When the different source domains hold different label sets, such learning causes mismatch of classes. CIDDG~\cite{cite:ECCV18CIAN} can avoid the mismatching but still requires all the source and target domains to share the same label sets, or otherwise the low domain diversity of some classes makes it hard to learn domain-invariant features. 

Meta-learning instead has the potential to learn from highly diverse domains. However, current meta-learning-based domain generalization methods still fail to consider different label sets of distinct source domains and the open classes in the target domain~\cite{cite:AAAI18MLDG,cite:NIPS18MetaReg,cite:NIPS19MASF,cite:ICCV19Epic}. Heterogeneous domain generalization~\cite{li2019feature,wang2020heterogeneous} has a similar goal of learning generalizable representations, which targets a more powerful pre-trained model by learning from heterogeneous source domains of different label sets. However, it requires additional target labeled data to induce a category model, which cannot fit into the proposed open domain generalization problem.

\textbf{Augmentation} The statistical learning theory~\cite{vapnik2013nature} suggests that the generalization of the learning model can be characterized by the model capacity and the diversity of training data. So data augmentation can improve generalization by increasing the diversity of training data. Basic augmentations including affine transformation, random cropping, and horizontal flipping are widely-used in image classification~\cite{ciregan2012multi,sato2015apac,krizhevsky2017imagenet}. Recently, more advanced augmentations are proposed. Mixup~\cite{zhang2017mixup,tokozume2018between,guo2019mixup} combines two samples linearly. Cutout~\cite{devries2017improved} removes contiguous sections of input images. Cutmix~\cite{yun2019cutmix} combines cutout and mixup by filling the Cutout part with sections of other image patches.  

Augmentation-based generalization methods promote the generalization ability by augmenting source data, where adversarial data augmentation~\cite{volpi2018generalizing}, gradient-based perturbations~\cite{cite:ICLR18CROSSGRAD}, self-supervised learning signals~\cite{cite:CVPR19JiGen}, and CutMix~\cite{cite:ECCV20Cumix} are used as the augmentation method. Note that these augmentation methods target general situations for generalization across domains but are not designed specially for open domains with disparate label sets. 

Different from all previous works, this paper studies open domain generalization, a practical but challenging problem.
We develop the DAML framework to conduct meta-learning over augmented source domains. We design a novel Dir-mixup to mix samples from multiple domains instead of mixing two arbitrary samples in classic mixup. Dir-mixup bridges all the source domains and compensates each domain with missing classes from other domains, which naturally fits the disparate source label sets. We further propose a new distilled soft-labeling to transfer knowledge across domains.

\begin{figure*}[htbp]
  \centering
  \includegraphics[width=0.85\textwidth]{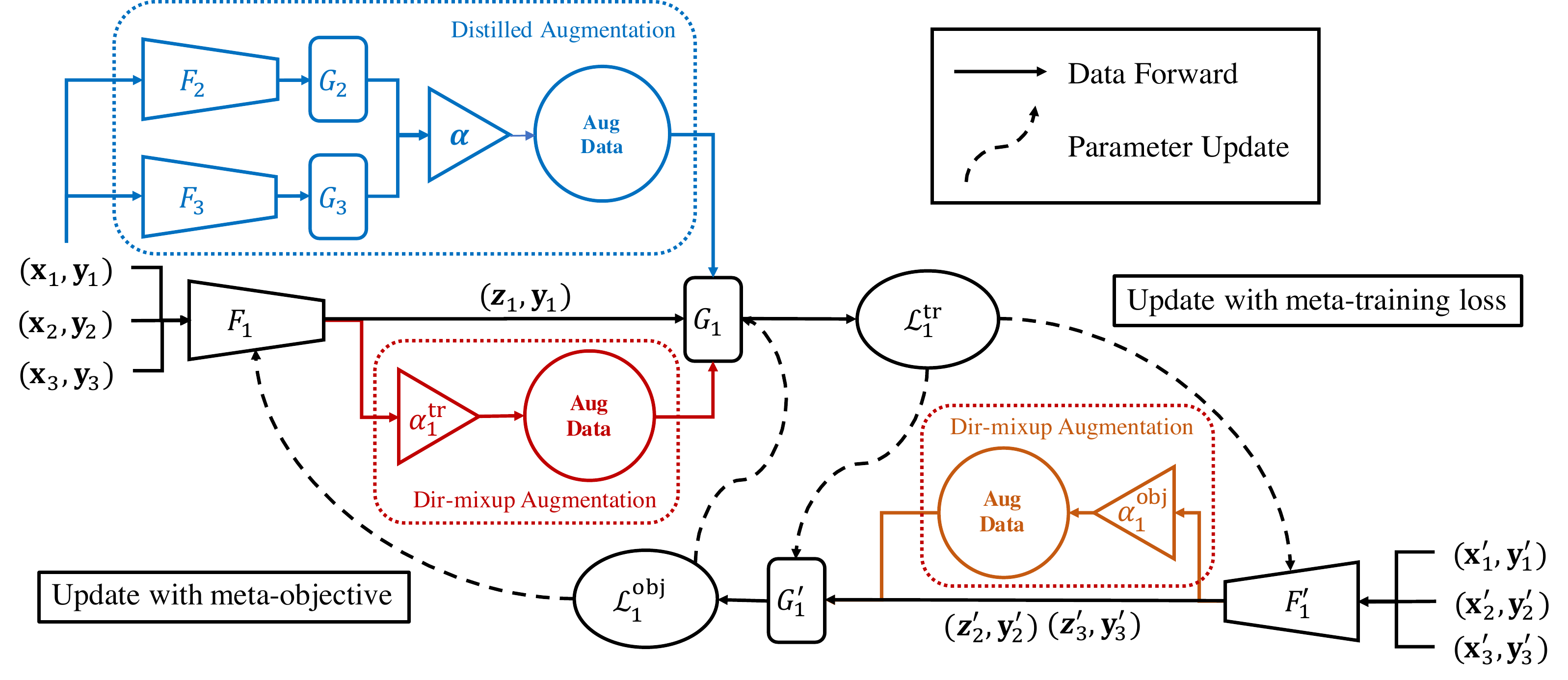}
  \vspace{-5pt}
  \caption{The architecture of the proposed DAML framework. We show the computation graph for source domain 1 as an example, and the other source domains are computed similarly. In the meta-training (up part, left to right), each source domain is augmented by Dir-mixup (red) and distilled soft-labeling (blue) to compute the $\mathcal{L}^\text{tr}_1$ to update the model parameters to $F^\prime_1$ and $G^\prime_1$. In the meta-objective (down part, right to left), each source domain is augmented by Dir-mixup (red) to compute the $\mathcal{L}^\text{obj}_1$ to finally update the model parameters.} 
   \label{fig:arch}
   \vspace{-10pt}
\end{figure*}

\section{Domain-Augmented Meta-Learning}
In this section, we first introduce the open domain generalization (OpenDG) problem. Then we introduce the Domain-Augmented Meta-Learning (DAML) and describe the step-by-step algorithm and the optimization of the framework, which consists of the proposed domain augmentation and the meta-learning on the augmented domains.

\subsection{Open Domain Generalization}
In open domain generalization (OpenDG), we have multiple source domains $\mathcal{D}_1,\mathcal{D}_2,\cdots,\mathcal{D}_S$ available for training, where each source domain $s$ consists of data-label pairs $\mathcal{D}_s=\{(\mathbf{x}_s,\mathbf{y}_s)\}$. $\mathbf{y}_s$ denotes the one-hot label of $\mathbf{x}_s$. Note that although we train the model with mini-batches in practice, here we omit the batch size of each domain to simplify the notations. We use $\mathcal{C}$ to denote the union of all the source label sets. In open domain generalization, we have no constraint on the label sets of different domains. We aim to learn open-domain generalizable representation from all the source domains, which can generalize well to an unseen target domain $\mathcal{D}_t$. Specifically, the target domain, only used for evaluation, consists of fully unlabeled data $\mathcal{D}_t=\{\mathbf{x}_t\}$ and its label set $\mathcal{C}_t$ may contain classes existing in any source label set or unknown classes not existing in the union of source label sets $\mathcal{C}$. The goal is to classify at inference each target sample with the correct class if it belongs to the source label set $\mathcal{C}$, or label it as ``unknown''. Note that no target data, even unlabeled, are available for training, which differs OpenDG from domain adaptation~\cite{you2019universal} or domain generalization~\cite{wang2020heterogeneous}.

\subsection{The DAML Framework}\label{sec:framework}
We propose DAML to address open domain generalization problems to mitigate the disparate label sets and distribution shifts among the diverse source domains. As shown in Algorithm~\ref{alg}, the idea is to learn generalizable representations by meta-learning over augmented domains.

\myparagraph{Augmented Domains}
As demonstrated in~\cite{zhang2016understanding,gong2019diversity}, increasing the diversity of the dataset can substantially improve the generalization of the representations. Motivated by this idea, we augment each domain to expand the diversity of the datasets. We observe that different domains have different distributions and hold different label sets, which means that each domain contains distinct knowledge but lacks domain knowledge and class knowledge of other domains. Based on the observation, we design domain augmentation to address open domain generalization. Our insight is to conduct both feature-level and label-level augmentation. For feature-level augmentation, we propose a novel Dirichlet Mixup (Dir-mixup) method, which augments each domain by the mixup with multiple domains. For label-level augmentation, we propose to augment each domain by distilling soft-labels from models of other domains. The proposed domain augmentation increases the diversity of the data and compensates each domain with missing knowledge of features and classes. The details of the proposed domain augmentation are introduced in Section~\ref{sec:domain_augmentation}.

\myparagraph{Meta-Learning}
We design the learning framework to learn generalizable representations, which simultaneously preserves the unique information of each domain and aggregates the knowledge of all the domains. Thus, instead of employing a shared network for all source domains, which only embeds domain common knowledge, we build one individual classification network composed of a feature extractor $F_s$ and a classifier $G_s$ for each source domain $s$. Then we need to learn a generalizable representation aggregating the information of all the source domains. We conduct meta-learning over all the networks since meta-learning is demonstrated to be able to learn a generalizable representation from highly disparate domains. In each iteration of the parameter update, we first draw a batch of samples from each domain and compute the corresponding Dir-mixup samples and distilled soft-labels (Line 5-7 in Algorithm~\ref{alg}). Unlike standard meta-learning loss applied only on the raw data~\cite{finn2017model}, with the augmented domains, we design a new meta-training loss as the classification loss on the original data, the domain-augmented data by Dir-mixup, and soft-labels distilled from other domain networks. For each domain $s$, let $\mathbf{z}_s = F_s(\mathbf{x}_s)$ be the feature of $\mathbf{x}_s$, we define the meta-training loss as
\begin{equation}\label{eqn:meta-train}
\small
\begin{aligned}
\mathcal{L}^{\text{tr}}_s&=\mathop{\mathbb{E}}_{(\mathbf{x}_s,\mathbf{y}_s)\sim \mathcal{D}_s}\left[-\sum_{k=1}^{|\mathcal{C}|}(\mathbf{y}_s)^{(k)}\log\left(G_s^{(k)} (F_s(\mathbf{x}_s))\right)\right] \\
    &+\mathop{\mathbb{E}}_{(\mathbf{z}^{\text{D-mix}}_s,\mathbf{y}^{\text{D-mix}}_s)\sim \mathcal{D}^{\text{D-mix}}_s}\left[-\sum_{k=1}^{|\mathcal{C}|}(\mathbf{y}_s^{\text{D-mix}})^{(k)}\log\left(G_s^{(k)}(\mathbf{z}^{\text{D-mix}}_s)\right)\right]\\
    &+\mathop{\mathbb{E}}_{(\mathbf{x}_s,\mathbf{y}_s^{\text{distill}})\sim \mathcal{D}_s^{\text{distill}}}\left[-\sum_{k=1}^{|\mathcal{C}|}(\mathbf{y}_{s}^{\text{distill}})^{(k)}\log\left({G_s^{(k)}(F_s(\mathbf{x}_s)})\right)\right].
\end{aligned}
\end{equation}

The superscript $(k)$ means the probability of the $k$-th class. $\mathcal{D}^{\text{D-mix}}_s$ and $\mathcal{D}^{\text{distill}}_s$ are the augmented domains of Dir-mixup samples and distilled soft-label samples for meta-training on domain $s$. We compute one step of gradient update for each source network with respect to the meta-training loss:  $\theta_{G^\prime_s,F^\prime_s} = \theta_{G_s,F_s}-\eta \nabla_\theta \mathcal{L}_{s}^\text{tr}$ (Line 9 in Algorithm~\ref{alg}), where $\eta$ is the step size. The design idea of meta-objective is to guide the gradient update from the meta-training loss to the desired goal. Classic meta-learning employs the losses over all sampled tasks as the meta-objective~\cite{finn2017model}. But our goal is to improve the generalization ability of the model, so different from classic meta-objective, we design the meta-objective as the classification loss on the original data and Dir-mixup data in other domains with the updated network $G'_s,F'_s$, which can propagate the knowledge of other domains to domain $s$ and promote the knowledge transfer and generalization across domains. The meta-objective is defined as 
\begin{equation}\label{eqn:meta-val}
\small
\begin{aligned}
&\mathcal{L}^{\text{obj}}_s = \sum_{j \neq s}\mathop{\mathbb{E}}_{(\mathbf{x}_j,\mathbf{y}_j)\sim \mathcal{D}_j}\left[-\sum_{k=1}^{|\mathcal{C}|}(\mathbf{y}_j)^{(k)}\log\left(G_s^{\prime{(k)}} (F_s^{\prime}(\mathbf{x}_j))\right)\right] \\
    &+\mathop{\mathbb{E}}_{(\mathbf{z}^{\text{D-mix}^{\prime}}_s,\mathbf{y}^{\text{D-mix}^{\prime}}_s)\sim \mathcal{D}^{\text{D-mix}^{\prime}}_s}\left[-\sum_{k=1}^{|\mathcal{C}|}(\mathbf{y}_s^{\text{D-mix}^{\prime}})^{(k)}\log\left(G_s^{{\prime}(k)}(\mathbf{z}^{\text{D-mix}^{\prime}}_s)\right)\right]
\end{aligned}
\end{equation}
${\mathcal{D}^{\text{D-mix}}_s}^\prime$ is the augmented domain of Dir-mixup samples for domain $s$ in meta-objective. The minimization of the meta-objective finds a gradient descent update that updates the network to classify data in other domains with high accuracy, which encourages the network to learn a generalizable representation performing well across all domains. We finally update the network parameters in one iteration by $\theta_{s} \leftarrow \theta_s-\beta \nabla_\theta (\mathcal{L}^{\text{tr}}_{s}+ \mathcal{L}_{s}^\text{obj})$, where $\beta$ is the learning rate.

\subsection{Domain Augmentation}\label{sec:domain_augmentation}
The meta-learning framework can learn a generalizable representation aggregating information from all source domains, where the generalization power highly relies on the diversity of each source domain. To this end, we design two multiple source domain augmentation approaches: the feature-level augmentation, Dir-mixup, and the label-level augmentation, distilled augmentation. The augmentations compensate for the missing class information in each source domain and further increase domain diversity.

\textbf{Dir-mixup} Mixup~\cite{zhang2017mixup} generates a new data-label by the weighted sum of the feature and one-hot label of existing samples, where the weights are sampled from a pre-defined distribution. We augment the $s$-th source domain by mixup of data in the $s$-th domain with data in other domains. Since these data may belong to the missing classes of the $s$-th source domain, mixup augmentation would compensate for the missing classes. Also, mixup produces inter-domain data, which further increases the diversity of data in each domain. 

\begin{figure}[htbp]
  \centering
  \vspace{-5pt}
  \includegraphics[width=0.45\textwidth]{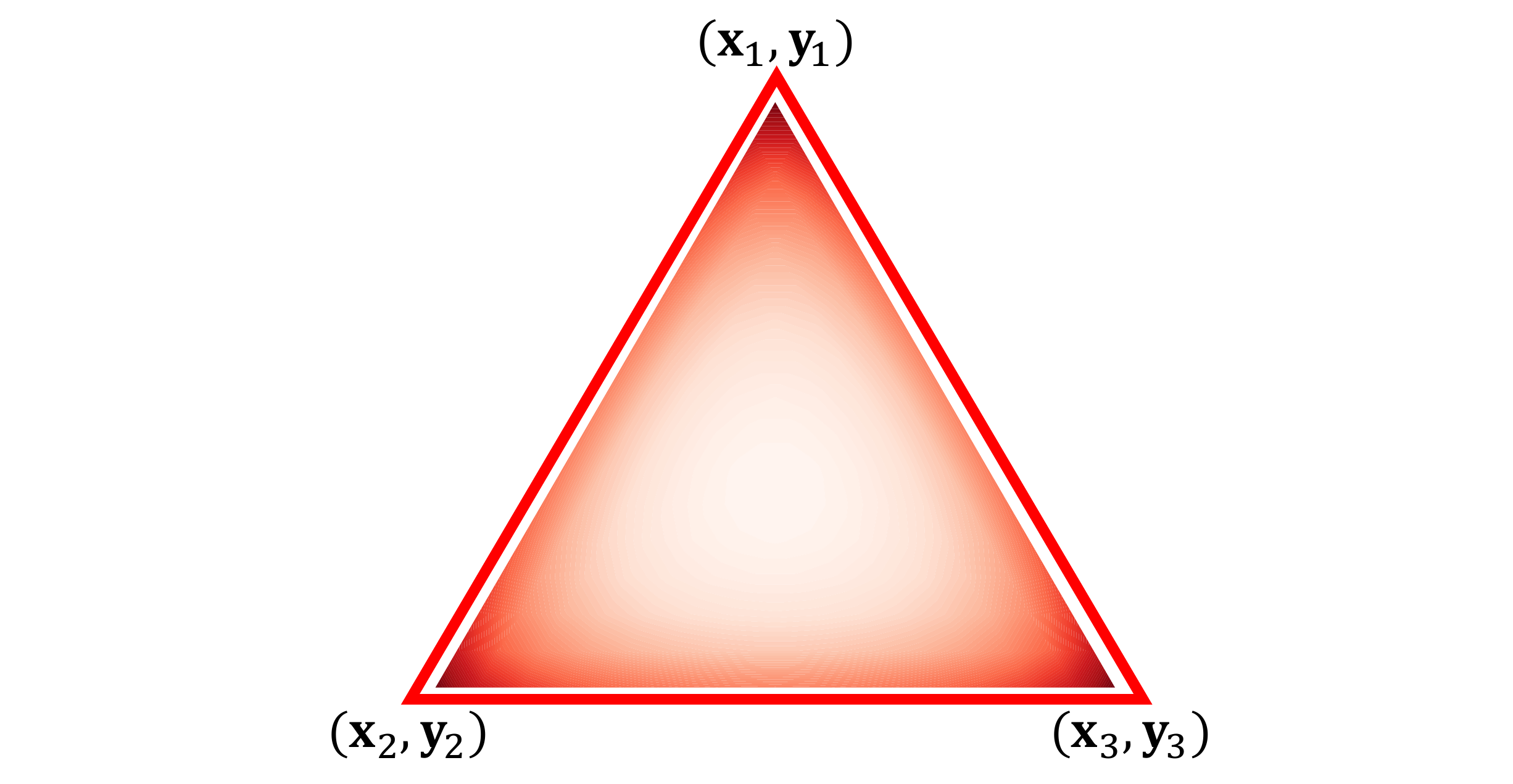}
  \vspace{-5pt}
  \caption{Comparison between Dir-mixup and classic mixup. Classic mixup only mixes two samples, so mixup samples only exist on the edge of the triangle while Dir-mixup mixes samples of multiple domains covering the whole triangle area, meaning Dir-mixup introduce mixup samples with more information and higher diversity.} 
   \label{fig:dirichlet}
   \vspace{-5pt}
\end{figure}

However, the original mixup is defined to mix two samples. When applied to open domain generalization with multiple source domains, mixup samples are only generated from pairs of domains, which, as shown in Figure~\ref{fig:dirichlet}, only generates samples between two domains (the lines between vertex) but lacks samples mixing multiple domains (the whole area). Also, to obtain all domain combinations, such pairwise mixup needs $O$(\#domains$\times$\#domains) mixup samples. Therefore, to mix multiple domains, we need to sample the weight from a multi-variate distribution instead of the beta distribution used in the original mixup.
We select Dirichlet distribution since it has similar properties to the beta distribution and is a multi-variate distribution. We then design a new Dir-mixup to mix samples (one for each domain) with a designed weight $\bm{\lambda}$ sampled from a Dirichlet distribution parameterized by a parameter $\bm{\alpha}$. We perform mixup at feature-level. Let $\mathbf{z}_1, \mathbf{z}_2, \cdots, \mathbf{z}_S$ be the features of different domain data extracted by the network, the Dir-mixup augmented data $(\mathbf{z}^{\text{D-mix}}, \mathbf{y}^{\text{D-mix}})$ can be calculated as:
\begin{equation}\label{eqn:mixup}
\begin{aligned}
    \bm{\lambda} &\sim \text{Dirichlet}(\bm{\alpha})\\
    (\mathbf{z}^{\text{D-mix}}, \mathbf{y}^{\text{D-mix}})  &= (\sum_{s=1}^{S} {\lambda^{(s)}}{\mathbf{z}_s}, \sum_{s=1}^{S} {\lambda^{(s)}}{\mathbf{y}_s}).
\end{aligned}
\end{equation}

Compared with recent work using mixup for domain generalization~\cite{cite:ECCV20Cumix, wang2020heterogeneous}, Dir-mixup is more efficient and effective. The parameter $\bm{\alpha}$ adjusts the distribution to generate different augmentations, better serving the meta-learning process. Consider constructing Dir-mixup for each model $s$. In the meta-training, we want to keep more information and focus more on domain $s$ during mixup, so we set $\alpha^{(s)}$ larger than other components in $\bm{\alpha}$, which assigns a larger weight $\lambda^{(s)}$ to $\mathbf{z}_s$ statistically. In the meta objective, the goal is to transfer knowledge from other domains and improve the cross-domain generalization, which would be enhanced by mixup results with larger domain discrepancy. So we set $\alpha^{(s)}$ smaller than other components in $\bm{\alpha}$, which induces smaller $\lambda^{(s)}$ statistically. We employ two hyper-parameters $\alpha_\text{max}$ and $\alpha_\text{min}$ to realize this idea. For the meta-training of model $s$, we set $\bm{\alpha}^{\text{tr}}_s$ to be a length $S$ vector with all entries as $\alpha_{\text{min}}$ but the $s$-th entry as $\alpha_{\text{max}}$. We generate mixup data with this $\bm{\alpha}^{\text{tr}}_s$ to form the Dir-mixup augmentation set in the meta-training of model $s$, as $\mathcal{D}^{\text{D-mix}}_s$ in Equation~\ref{eqn:meta-train}. For the meta-objective, we set $\bm{\alpha}_s^{\text{obj}}$ to be a length $S$ vector with all entries as $\alpha_{\text{max}}$ but the $s$-th entry as $\alpha_{\text{min}}$. And the data generated from this $\bm{\alpha}^{\text{obj}}_s$ form the Dir-mixup augmentation set for model $s$, which is the $\mathcal{D}^{\text{D-mix}^{\prime}}_s$ in Equation~\ref{eqn:meta-val}.

\textbf{Distilled Augmentation}
For the $s$-th source domain, we further augment it with the soft-labeling distilled from other domains, which is the output predictions of other networks. We mix soft-labels from other domains to increase the diversity of the augmentation. We set the $\bm{\alpha}$ to be a vector of all ones with dimension $S-1$ since we do not prefer a particular other domain. The augmentation can be defined as
\begin{equation}\label{eqn:distill}
\begin{aligned}
    \bm{\lambda} &\sim \text{Dirichlet}(\bm{\alpha})\\
   \mathbf{y}_{s}^{\text{distill}} =  \sum_{j=1}^{s-1}\lambda^{(j)}G_j(&F_j(\mathbf{x}_s))+\sum_{j=s+1}^{S}\lambda^{(j-1)}G_j(F_j(\mathbf{x}_s)).
\end{aligned}
\end{equation}
The soft-label indicates the decision of the networks of other domains on the $s$-th domain data, which transfers the knowledge from other domains to the $s$-th domain. The augmentation is reflected as the third term in Equation~\ref{eqn:meta-train}, where we do not back-propagate through $F_j$, $G_j$ since they are just used to generate the soft-labeling. The augmentation regularizes the $s$-th domain network with knowledge of other domains, which derives a more generalizable representation.

\subsection{Inference}
In the inference stage, we have the networks for all source domains $G_1,\cdots,G_S,F_1,\cdots,F_S$ trained by the DAML framework as shown in Algorithm~\ref{alg}. For a test sample $\mathbf{x}_t$ from the target domain $\mathcal{D}_t$, we compute the raw prediction of $\mathbf{x}_t$ by aggregating the predictions of all the source networks:
\begin{equation}
    \widehat{\mathbf{y}}_t = \frac{1}{S}\sum_{s=1}^{S}G_s(F_s(\mathbf{x}_t)).
\end{equation}
The ensemble of all source domain networks naturally calibrates the prediction confidence and enables DAML to achieve higher performance in the unseen target domain.

\begin{table*}[ht]
	\begin{center}
		\caption{Results of PACS dataset under the open-domain setting.}
		\label{table:PACS}
		\resizebox{0.98\textwidth}{!}{%
				\begin{tabular}{lcccccccccc}
					\hline
					&\multicolumn{2}{c}{\textbf{Art}}&\multicolumn{2}{c}{\textbf{Sketch}}&\multicolumn{2}{c}{\textbf{Photo}}&\multicolumn{2}{c}{\textbf{Cartoon}}&\multicolumn{2}{c}{\textbf{Avg}} \\
					\textbf{Method}&Acc&H-score&Acc&H-score&Acc&H-score&Acc&H-score&Acc&H-score \\
					\hline
					AGG &$51.35$&$38.87$&$49.75$&$47.09$&$53.15$&$44.19$&$66.43$&$48.98$&$55.17\pm{0.16}$&$44.78\pm{0.33}$\\
					MLDG~\cite{cite:AAAI18MLDG} &$44.59$&$31.54$&$51.29$&$49.91$&$62.20$&$43.35$&$71.64$&$55.20$&$57.43\pm{0.14}$&$45.00\pm{0.31}$\\
					FC~\cite{li2019feature} &$51.12$&$39.01$&$51.15$&$49.28$&$60.94$&$45.79$&$69.32$&$52.67$&$58.13\pm{0.20}$&$46.69\pm{0.25}$\\
					Epi-FCR~\cite{cite:ICCV19Epic} &$\mathbf{54.16}$&$41.16$&$46.35$&$46.14$&$70.03$&$48.38$&$72.00$&$\mathbf{58.19}$&$60.64\pm{0.22}$&$48.47\pm{0.29}$\\
					PAR~\cite{cite:NIPS19PAR} &$52.97$&$39.21$&$53.62$&$52.00$&$51.86$&$36.53$&$67.77$&$52.05$&$56.56\pm{0.51}$&$44.95\pm{0.57}$\\
					RSC~\cite{cite:ECCV20RSC} &$50.47$&$38.43$&$50.17$&$44.59$&$67.53$&$49.82$&$67.51$&$47.35$&$58.92\pm{0.46}$&$45.05\pm{0.60}$\\
					CuMix~\cite{cite:ECCV20Cumix} &$53.85$&$38.67$&$37.70$&$28.71$&$65.67$&$49.28$&$\mathbf{74.16}$&$47.53$&$57.85\pm{0.32}$&$41.05\pm{0.66}$\\
					\hline
					DAML (ours) &$54.10$&$\mathbf{43.02}$&$\mathbf{58.50}$&$\mathbf{56.73}$&$\mathbf{75.69}$&$\mathbf{53.29}$&$73.65$&$54.47$&$\mathbf{65.49}\pm{0.36}$&$\mathbf{51.88}\pm{0.42}$\\
					\hline
				\end{tabular}%
		}
	\end{center}
\end{table*}

\begin{table*}[ht]
	\begin{center}
		\caption{Results of Office-Home dataset under the open-domain setting.}
		\label{table:OfficeHome}
		\resizebox{0.98\textwidth}{!}{%
				\begin{tabular}{lcccccccccc}
					\hline
					&\multicolumn{2}{c}{\textbf{Clipart}}&\multicolumn{2}{c}{\textbf{Real-World}}&\multicolumn{2}{c}{\textbf{Product}}&\multicolumn{2}{c}{\textbf{Art}}&\multicolumn{2}{c}{\textbf{Avg}} \\
					\textbf{Method}&Acc&H-score&Acc&H-score&Acc&H-score&Acc&H-score&Acc&H-score \\
					\hline
					AGG &$42.83$&$\mathbf{44.98}$&$62.40$&$53.67$&$54.27$&$50.11$&$42.22$&$40.87$&$50.43\pm{0.32}$&$47.41\pm{0.53}$\\
					MLDG~\cite{cite:AAAI18MLDG} &$41.82$&$41.26$&$62.98$&$55.84$&$56.89$&$52.25$&$42.58$&$40.97$&$51.07\pm{0.19}$&$47.58\pm{0.42}$\\
					FC~\cite{li2019feature} &$41.80$&$41.65$&$63.79$&$55.16$&$54.41$&$52.02$&$44.13$&$43.25$&$51.03\pm{0.24}$&$48.02\pm{0.57}$\\
					Epi-FCR~\cite{cite:ICCV19Epic} &$37.13$&$42.05$&$62.60$&$54.73$&$54.95$&$52.68$&$46.33$&$44.46$&$50.25\pm{0.50}$&$48.48\pm{0.76}$\\
					PAR~\cite{cite:NIPS19PAR} &$41.27$&$41.77$&$65.98$&$57.60$&$55.37$&$54.13$&$42.40$&$42.62$&$51.26\pm{0.27}$&$49.03\pm{0.41}$\\
				    RSC~\cite{cite:ECCV20RSC} &$38.60$&$38.39$&$60.85$&$53.73$&$54.61$&$54.66$&$44.19$&$44.77$&$49.56\pm{0.44}$&$47.89\pm{0.79}$\\
					CuMix~\cite{cite:ECCV20Cumix} &$41.54$&$43.07$&$64.63$&$58.02$&$57.74$&$55.79$&$42.76$&$40.72$&$51.67\pm{0.12}$&$49.40\pm{0.27}$\\
					\hline
					DAML (ours) &$\mathbf{45.13}$&$43.12$&$\mathbf{65.99}$&$\mathbf{60.13}$&$\mathbf{61.54}$&$\mathbf{59.00}$&$\mathbf{53.13}$&$\mathbf{51.11}$&$\mathbf{56.45}\pm{0.21}$&$\mathbf{53.34}\pm{0.45}$\\
					\hline
				\end{tabular}%
		}
	\end{center}
\end{table*}

\section{Experiments}
We construct several open domain generalization scenarios with different datasets to evaluate the proposed method. 
\subsection{Datasets}
\textbf{PACS} dataset~\cite{cite:ICCV17DBA} consists of four domains corresponding to four different image styles, including photo (\textbf{P}), art-painting (\textbf{A}), cartoon (\textbf{C}) and sketch (\textbf{S}). The four domains have the same label set of 7 classes. We use each domain as the target domain and the other three domains as source domains to form four cross-domain tasks. We evaluate the generalization performance on both the original closed-set dataset and the modified open-domain dataset.

\textbf{Office-Home}~\cite{cite:CVPR17OfficeHome} comprises of images from four different domains: Artistic (\textbf{Ar}), Clip art (\textbf{Cl}), Product (\textbf{Pr}) and Real-world (\textbf{Rw}). It has a large domain gap and 65 classes which is much more than other DG datasets, so it is very challenging. We spread these 65 classes among the four domains to derive an open-domain dataset. We construct four open generalization tasks based on it, where each domain is used as the target domain respectively, and the other three domains serve as source domains. 

\textbf{Multi-Datasets} scenario is constructed in this paper to consider a more realistic situation of learning generalizable representations from arbitrary source domains. We simulate the process where we obtain source domains from different resources and try to learn a generalizable model to achieve high accuracy on an unseen target domain. We leverage several public datasets including \textbf{Office-31}~\cite{cite:ECCV10Office}, \textbf{STL-10}~\cite{coates2011analysis} and \textbf{Visda2017}~\cite{visda2017} as source domains, and evaluate the generalization performance on four domains in \textbf{DomainNet}~\cite{peng2019moment}. There exist distribution discrepancy and huge label-set disparity across the four datasets, which forms a natural open domain generalization scenario. Since there are too many open classes in the DomainNet, we preserve all the classes existing in the joint label set of source domains and subsample 20 open classes.

\begin{table}[htbp]
    \addtolength{\tabcolsep}{1pt} 
	\begin{center}
		\caption{Results on closed-set PACS dataset.}
		\label{table:PACS2}
		\resizebox{0.99\columnwidth}{!}{%
				\begin{tabular}{lccccc}
					\hline
					\textbf{Method}&\textbf{A}&\textbf{S}&\textbf{P}&\textbf{C}&\textbf{Avg} \\
					\hline
					AGG &$77.6$&$70.3$&$94.4$&$73.9$&$79.1$\\
					CIDDG~\cite{cite:ECCV18CIAN} &$82.0$&$74.8$&$94.6$&$74.4$&$81.4$\\
					MLDG~\cite{cite:AAAI18MLDG} &$79.5$&$71.5$&$94.3$&$77.3$&$80.7$\\
					CrossGrad~\cite{cite:ICLR18CROSSGRAD} &$78.7$&$65.1$&$94.0$&$73.3$&$77.8$\\
					MetaReg~\cite{cite:NIPS18MetaReg} &$79.5$&$72.2$&$94.3$&$75.4$&$80.4$\\
					JiGen~\cite{cite:CVPR19JiGen} &$79.4$&$71.4$&$\mathbf{96.0}$&$75.3$&$80.4$\\
					MASF~\cite{cite:NIPS19MASF} &$80.3$&$71.7$&$94.5$&$77.2$&$81.0$\\
					Epi-FCR~\cite{cite:ICCV19Epic} &$82.1$&$73.0$&$93.9$&$77.0$&$81.5$\\
					CSD~\cite{cite:ICML20CSD} &$79.8$&$72.5$&$95.5$&$75.0$&$80.7$\\
					DMG~\cite{cite:ECCV20DMG} &$76.9$&$\mathbf{75.2}$&$93.4$&$\mathbf{80.4}$&$81.5$\\
					CuMix~\cite{cite:ECCV20Cumix} &$82.3$&$72.6$&$95.1$&$76.5$&$81.6$\\
					\hline
					DAML &$\mathbf{83.0}$&$74.1$&$95.6$&$78.1$&$\mathbf{82.7}$\\
					\hline
				\end{tabular}%
		}
	\end{center}
	\vspace{-10pt}
\end{table}

\subsection{Closed-Set Generalization}
We evaluate the classification accuracy of closed-set generalization on the widely-used domain generalization dataset \textbf{PACS}. The closed-set setting exactly matches the domain generalization setting so we compare with supervised learning on the merged datasets of all source domains: AGG, domain generalization methods including domain-invariant feature learning based methods: CIDDG~\cite{cite:ECCV18CIAN}, CSD~\cite{cite:ICML20CSD} and DMG~\cite{cite:ECCV20DMG}, meta-learning based methods: MLDG~\cite{cite:AAAI18MLDG}, MetaReg~\cite{cite:NIPS18MetaReg}, MASF~\cite{cite:NIPS19MASF} and Epi-FCR~\cite{cite:ICCV19Epic}, and augmentation based methods: CrossGrad~\cite{cite:ICLR18CROSSGRAD}, JiGen~\cite{cite:CVPR19JiGen} and CuMix~\cite{cite:ECCV20Cumix}. We do not compare with domain adaptation methods since they need unlabeled target data.

As shown in Table~\ref{table:PACS2}, on the closed-set generalization setting, to which previous domain generalization methods are tailored, DAML still outperforms all previous methods on average and achieves at least comparable performance on all the tasks. In particular, DAML outperforms state-of-the-art meta-learning-based DG, which indicates the importance of domain augmentation to learn generalizable representations. DAML surpasses state-of-the-art augmentation-based DG, indicating that the meta-learning paradigm and the carefully designed feature-level and label-level augmentations can enable learning more generalizable representations. 

\begin{table*}[ht]
	\begin{center}
		\caption{Results on the Multi-Datasets scenario (naturally under the open-domain setting).}
		\label{table:MultiDataset}
		\resizebox{0.98\textwidth}{!}{%
				\begin{tabular}{lcccccccccc}
					\hline
					&\multicolumn{2}{c}{\textbf{Clipart}}&\multicolumn{2}{c}{\textbf{Real}}&\multicolumn{2}{c}{\textbf{Painting}}&\multicolumn{2}{c}{\textbf{Sketch}}&\multicolumn{2}{c}{\textbf{Avg}} \\
					\textbf{Method}&Acc&H-score&Acc&H-score&Acc&H-score&Acc&H-score&Acc&H-score \\
					\hline
					AGG &$29.78$&$34.06$&$65.33$&$64.72$&$44.30$&$51.04$&$27.59$&$35.41$&$41.75\pm{0.63}$&$46.31\pm{0.57}$\\
					MLDG~\cite{cite:AAAI18MLDG} &$29.66$&$35.11$&$65.37$&$54.40$&$44.04$&$50.53$&$26.83$&$34.57$&$41.48\pm{0.68}$&$43.65\pm{0.71}$\\
					FC~\cite{li2019feature} &$29.91$&$35.42$&$64.77$&$63.65$&$44.13$&$50.07$&$28.56$&$34.10$&$41.84\pm{0.73}$&$45.81\pm{0.69}$\\
					Epi-FCR~\cite{cite:ICCV19Epic} &$27.70$&$37.62$&$60.31$&$64.95$&$39.57$&$50.24$&$26.76$&$33.74$&$38.59\pm{1.13}$&$46.64\pm{0.95}$\\
					PAR~\cite{cite:NIPS19PAR} &$29.29$&$39.99$&$64.09$&$62.59$&$42.36$&$46.37$&$30.21$&$39.96$&$41.49\pm{0.63}$&$47.23\pm{0.55}$\\
					RSC~\cite{cite:ECCV20RSC} &$27.57$&$34.98$&$60.36$&$60.02$&$37.76$&$42.21$&$26.21$&$30.44$&$37.98\pm{0.77}$&$41.91\pm{1.28}$\\
					CuMix~\cite{cite:ECCV20Cumix} &$30.03$&$40.18$&$64.61$&$65.07$&$44.37$&$48.70$&$29.72$&$33.70$&$42.18\pm{0.45}$&$46.91\pm{0.40}$\\
					\hline
					DAML (ours) &$\mathbf{37.62}$&$\mathbf{44.27}$&$\mathbf{66.54}$&$\mathbf{67.80}$&$\mathbf{47.80}$&$\mathbf{52.93}$&$\mathbf{34.48}$&$\mathbf{41.82}$&$\mathbf{46.61}\pm{0.59}$&$\mathbf{51.71}\pm{0.52}$\\
					\hline
				\end{tabular}%
		}
	\end{center}
	\vspace{-10pt}
\end{table*}

\begin{figure*}[ht]
    \centering
    \subfigure[Ar as the Target]{\includegraphics[width=.23\textwidth]{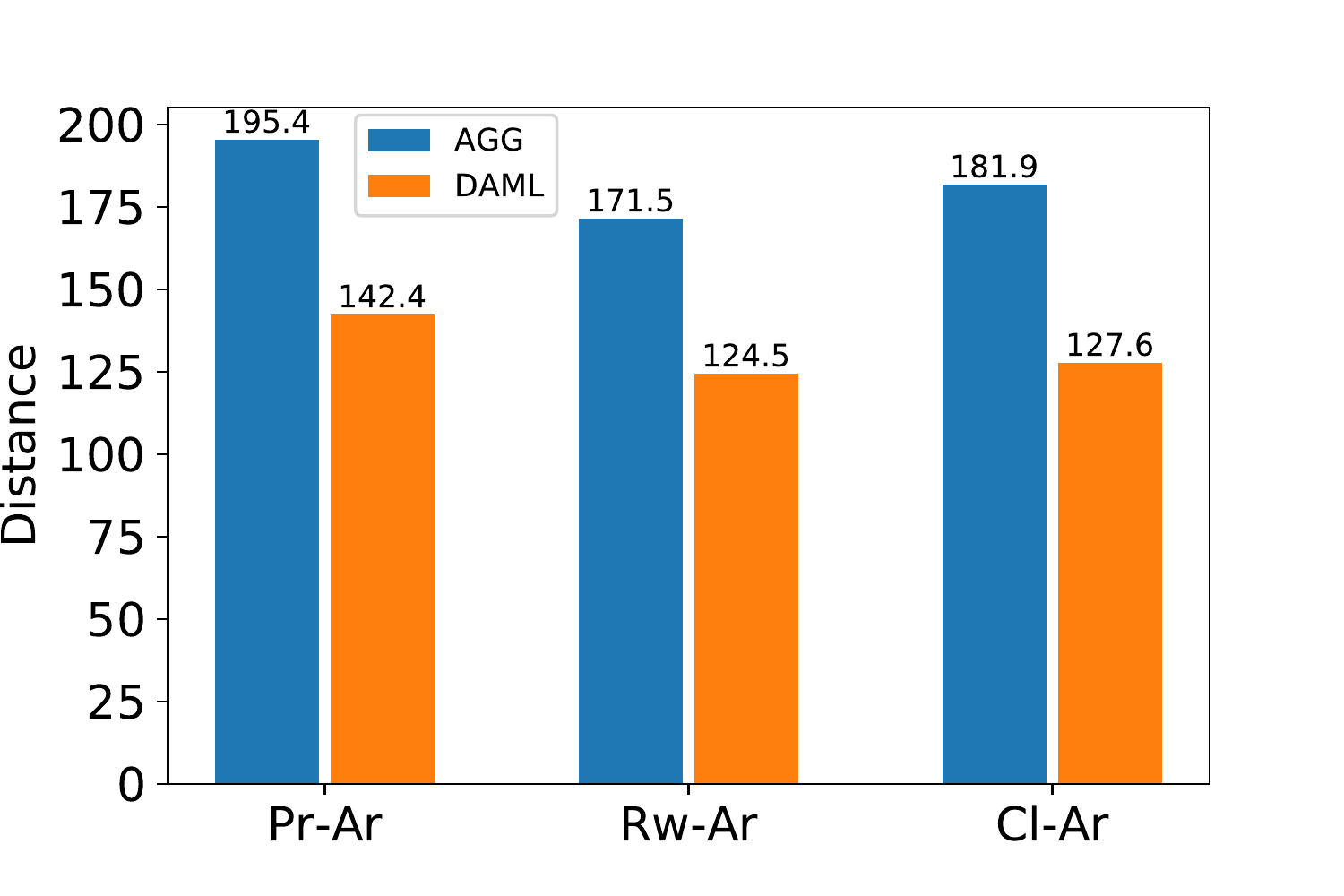}\label{fig:frechet_ar}}
    \subfigure[Cl as the Target]{\includegraphics[width=.23\textwidth]{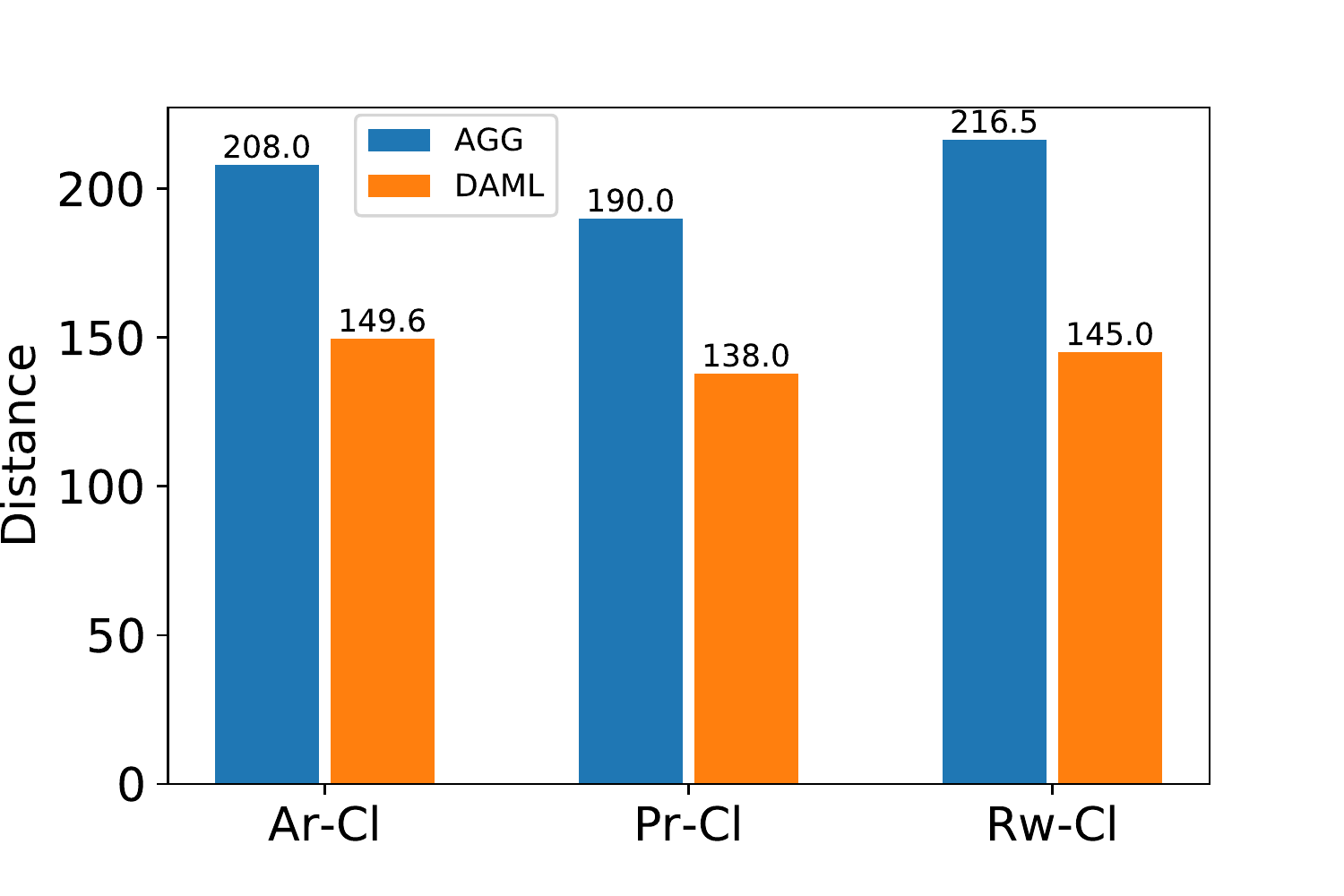}\label{fig:frechet_cl}}
    \subfigure[Pr as the Target]{\includegraphics[width=.23\textwidth]{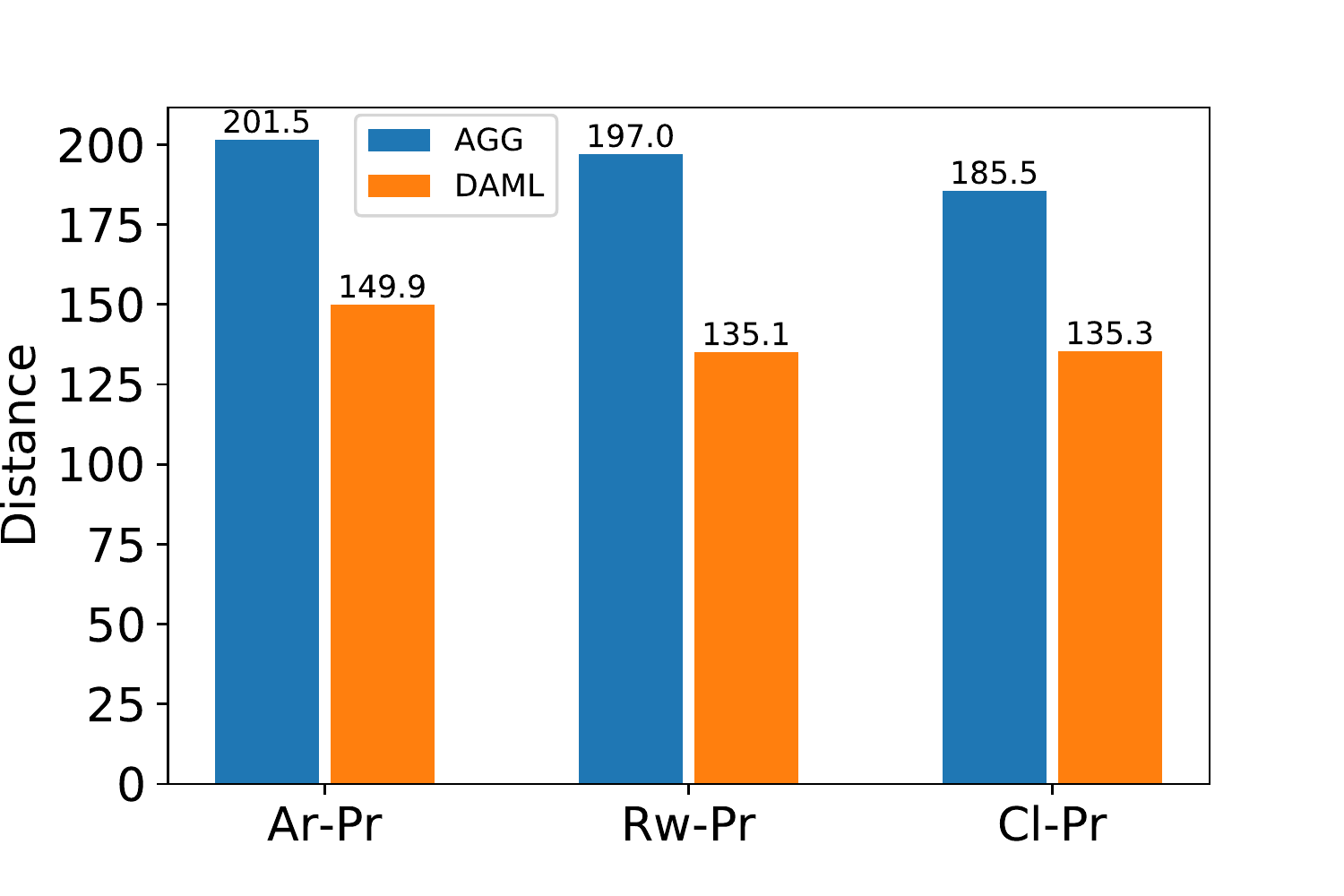}\label{fig:frechet_pr}}
    \subfigure[Rw as the Target]{\includegraphics[width=.23\textwidth]{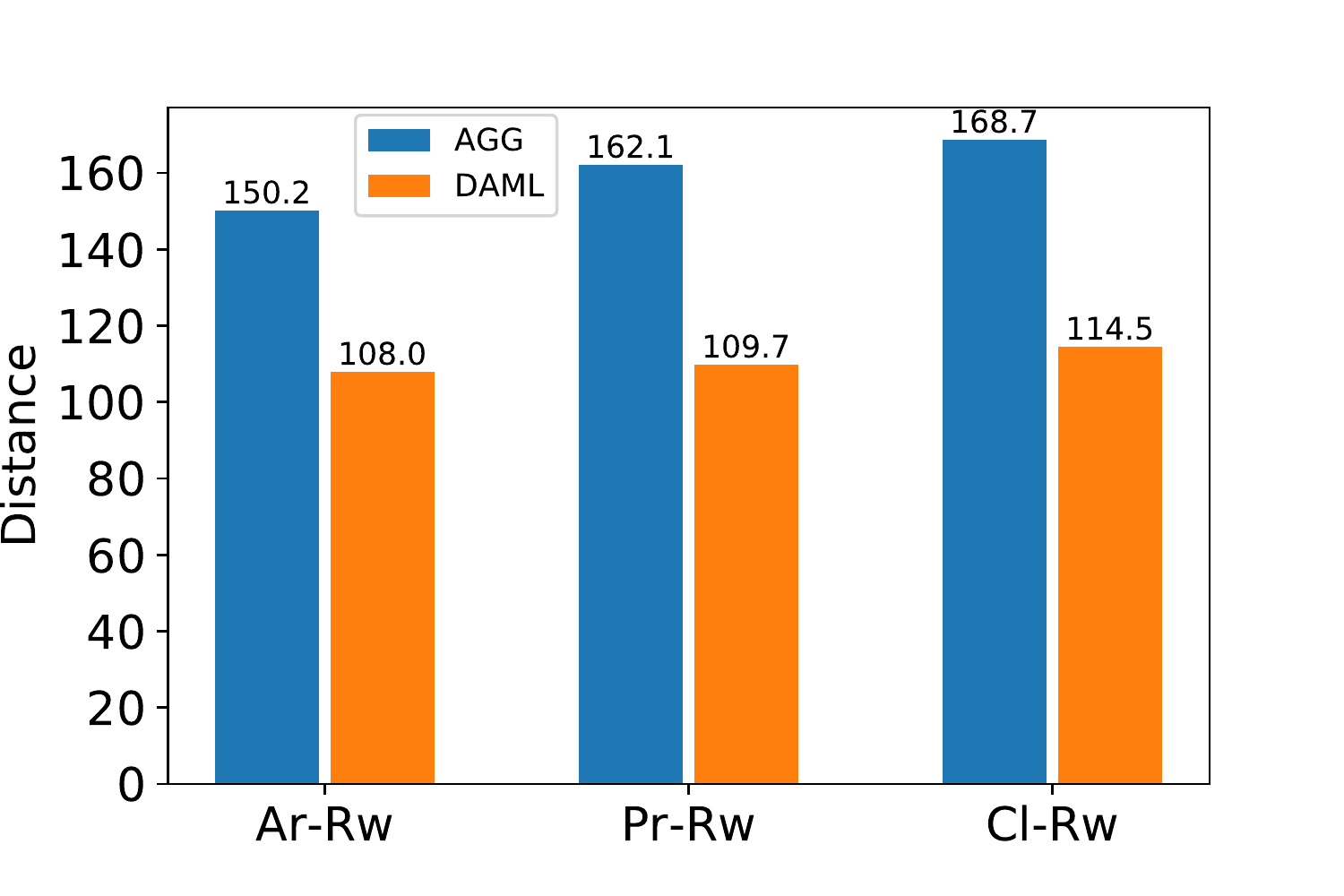}\label{fig:frechet_rw}}
    \caption{The Fr\'echet distance between each source domain and the target domain for the four generalization tasks on Office-Home dataset.}
    \label{fig:frechet_distance}
\end{figure*}

\subsection{Open Domain Generalization}
We evaluate the generalization performance for situations where the source and target domains have different label sets and open classes exist. We conduct experiments on PACS, Office-Home, and Multi-Datasets. For PACS and Office-Home, we preserve different parts of classes in the source domains and the target domain to create disparate label sets among source domains and between the source and target domains. For Multi-Datasets, we preserve all the classes for all source datasets. We show the class split in each domain in the supplementary materials. We follow~\cite{you2019universal} to set a threshold on the prediction confidence and label samples with a confidence lower than the threshold as an open class: ``unknown''. For the evaluation metric, we report the accuracy of data from non-open classes (Acc) and also follow the state-of-the-art universal domain adaptation paper~\cite{Fu_2020_ECCV} to use H-score to evaluate performance over all target data.

For the open-domain classification setting, we mainly compare with previous methods that are less influenced by the different label sets of source domains. We select state-of-the-art meta-learning-based and augmentation-based DG methods~\cite{cite:AAAI18MLDG,cite:ICCV19Epic,cite:ECCV20Cumix}, heterogeneous domain generalization methods: FC~\cite{li2019feature}, recently proposed methods of learning robust and generalizable features: PAR~\cite{cite:NIPS19PAR} and RSC~\cite{cite:ECCV20RSC}.

As shown in Tables~\ref{table:PACS},~\ref{table:OfficeHome} and~\ref{table:MultiDataset}, we can observe that DAML outperforms all the compared methods with a large margin on both Acc and H-score, which indicates that DAML not only learns a generalizable representation for non-open classes but also detects open classes with higher accuracy. In particular, DAML outperforms the meta-learning-based DG methods MLDG and Epi-FCR on almost all the tasks, especially the H-score, which demonstrates that domain augmentation, compensating missing labels for each domain, is vital to addressing the different label sets across source domains. DAML outperforms CuMix, which also employs mixup for data augmentation. Note that we design the Dir-mixup to mix samples from multiple domains while CuMix mixes two arbitrary samples. So our Dir-mixup creates mixup samples with higher variations and diversity, which encourages the model to learn more generalizable representations. 

The Multi-Datasets simulates the real-world scenario where we aim to generalize from datasets available at hand to an unseen domain. The different source domains hold extremely disparate label sets. In this realistic scenario, DAML outperforms all the compared methods with a large margin, indicating that DAML can be applied to realistic generalization problems and achieve higher performance.

\begin{table}[htbp]
    \addtolength{\tabcolsep}{-3pt} 
	\begin{center}
		\caption{Ablation study on the open-domain Office-Home dataset.}
		\label{table:Ablation}
		\resizebox{0.99\columnwidth}{!}{%
				\begin{tabular}{cccccccccc}
					\hline
					$\mathcal{D}^{\text{D-mix}}_s$&$\mathcal{D}^{\text{D-mix}^{\prime}}_s$&$\mathcal{D}^{\text{mix}}_s$&$\mathcal{D}^{\text{distill}}_s$& w/ Meta&\textbf{Cl}&\textbf{Rw}&\textbf{Pr}&\textbf{Ar}&\textbf{Avg} \\
					\hline
					-&-&-&-&\ymark&$42.2$&$64.8$&$57.6$&$49.6$&$53.6$\\
					\ymark&-&-&-&\ymark&$43.8$&$64.9$&$57.1$&$51.7$&$54.4$\\
					-&\ymark&-&-&\ymark&$43.8$&$65.7$&$58.2$&$52.4$&$55.0$\\
					\ymark&\ymark&-&-&\ymark&$44.8$&$65.9$&$59.7$&$52.9$&$55.9$\\
					\ymark&\ymark&-&\ymark&-&$44.1$&$65.1$&$59.7$&$52.2$&$55.3$\\
					-&-&\ymark&\ymark&\ymark&$44.3$&$65.3$&$59.0$&$51.9$&$55.1$\\
					\ymark&\ymark&-&\ymark&\ymark&$\mathbf{45.1}$&$\mathbf{66.0}$&$\mathbf{61.5}$&$\mathbf{53.1}$&$\mathbf{56.5}$\\
					\hline
				\end{tabular}%
		}
	\end{center}
	\vspace{-10pt}
\end{table}

\subsection{Analysis}
\textbf{Ablation Study} We go deeper into the DAML framework to explore the efficacy of each module in DAML including meta-learning, Dir-mixup and distilled soft-labels. As shown in Table~\ref{table:Ablation}, $\mathcal{D}^{\text{D-mix}}_s$ means whether to use the Dir-mixup data in the meta-training loss, \textit{i.e.} whether to use the second term in Equation~\ref{eqn:meta-train}. ${\mathcal{D}^{\text{D-mix}}_s}^\prime$ means whether to use the Dir-mixup data in the meta-objective loss, \textit{i.e.} whether to use the second term in Equation~\ref{eqn:meta-val}. $\mathcal{D}^{\text{mix}}_s$ means using classic mixup which mixes two arbitrary samples. $\mathcal{D}^{\text{distill}}_s$ means whether to use the distilled soft-label, \textit{i.e.} whether to use the third term in Equation~\ref{eqn:meta-train}. w/ Meta means whether to use meta-learning or otherwise supervised learning on the augmented domains.

In Table~\ref{table:Ablation}, we observe that using both $\mathcal{D}^{\text{D-mix}}_s$ and ${\mathcal{D}^{\text{D-mix}}_s}^\prime$ outperforms using only $\mathcal{D}^{\text{D-mix}}_s$ and using only ${\mathcal{D}^{\text{D-mix}}_s}^\prime$, which indicates Dir-mixup samples are helpful in both meta-training and meta-objective losses. Changing the Dir-mixup to classic mixup drops the accuracy, which shows the importance of a built-in mixup for multiple domains. Using $\mathcal{D}^{\text{distill}}_s$ outperforms not using $\mathcal{D}^{\text{distill}}_s$ on average, indicating that transferring knowledge between domains by distilled soft-labels learns more generalizable representations. DAML outperforms meta-learning conducted on the raw domain without any domain augmentation, which indicates the importance of domain augmentation to address the different label sets of source domains. DAML also outperforms the variant that uses no meta-learning, which demonstrates that meta-learning can aggregate knowledge from augmented source domains in a more effective way.

\textbf{Fr\'echet Distance} We compare the domain gap between source and target domains on features learned by the baseline AGG model and features learned by the DAML model. We extract features of each domain and compute their mean vectors and covariance matrices. Then we evaluate the Fr\'echet Distance\cite{dowson1982frechet} between the features of each source domain and the non-open class part of the target domain. As shown in Figure~\ref{fig:frechet_distance}, the domain gaps between source domains and the unseen target domain are smaller in DAML, indicating that DAML learns more generalizable representations.

\section{Conclusion}
In this paper, we propose a new open domain generalization problem aiming to generalize from arbitrary source domains with disparate label sets to unseen target domains, which can be widely utilized in real-world applications. We further propose a novel Domain-Augmented Meta-Learning framework (DAML) to address the problem, which conducts meta-learning over domains augmented at feature-level by specially designed Dir-mixup and at label-level by distilled soft-labels. Extensive experiments demonstrate that DAML learns more generalizable representations for classification in the target domain than the previous generalization methods.

{\small
\bibliographystyle{ieee_fullname}
\bibliography{egbib}
}

\appendix

\section{Experiment Details}
In this section, we clarify more details of the experiment settings due to the space limit in the main text.
\subsection{Datasets}
For each dataset, we show the exact class splits for each domain.

\textbf{PACS}~\cite{cite:ICCV17DBA} dataset consists of four domains corresponding to four different image styles, including photo (\textbf{P}), art-painting (\textbf{A}), cartoon (\textbf{C}) and sketch (\textbf{S}). The four domains have the same label set of 7 classes. We assign an index to each category, 0-Dog, 1-Elephant, 2-Giraffe, 3-Guitar, 4-Horse, 5-House, 6-Person. We use each domain as the target domain and the other three domains as source domains to form four cross-domain tasks: CPS-A, PAC-S, ACS-P, SPA-C. To construct the open-domain situations, we split the label space of the dataset, resulting in various label spaces across different domains. The specific categories contained in each domain are shown in Table \ref{table:PACSSupp}.

\begin{table}[htbp]
	\begin{center}
		\caption{Open-domain split of PACS dataset.}
		\label{table:PACSSupp}
				\begin{tabular}{cc}
					\hline
					\textbf{Domain}&\textbf{Classes} \\
					\hline
					Source-1 &$3, 0, 1$\\
					Source-2 &$4, 0, 2$\\
					Source-3 &$5, 1, 2$\\
					Target &$0, 1, 2, 3, 4, 5, 6$\\
					\hline
				\end{tabular}%
	\end{center}
	\vspace{-10pt}
\end{table}

\begin{figure}[htbp]
  \centering
  \includegraphics[width=0.45\textwidth]{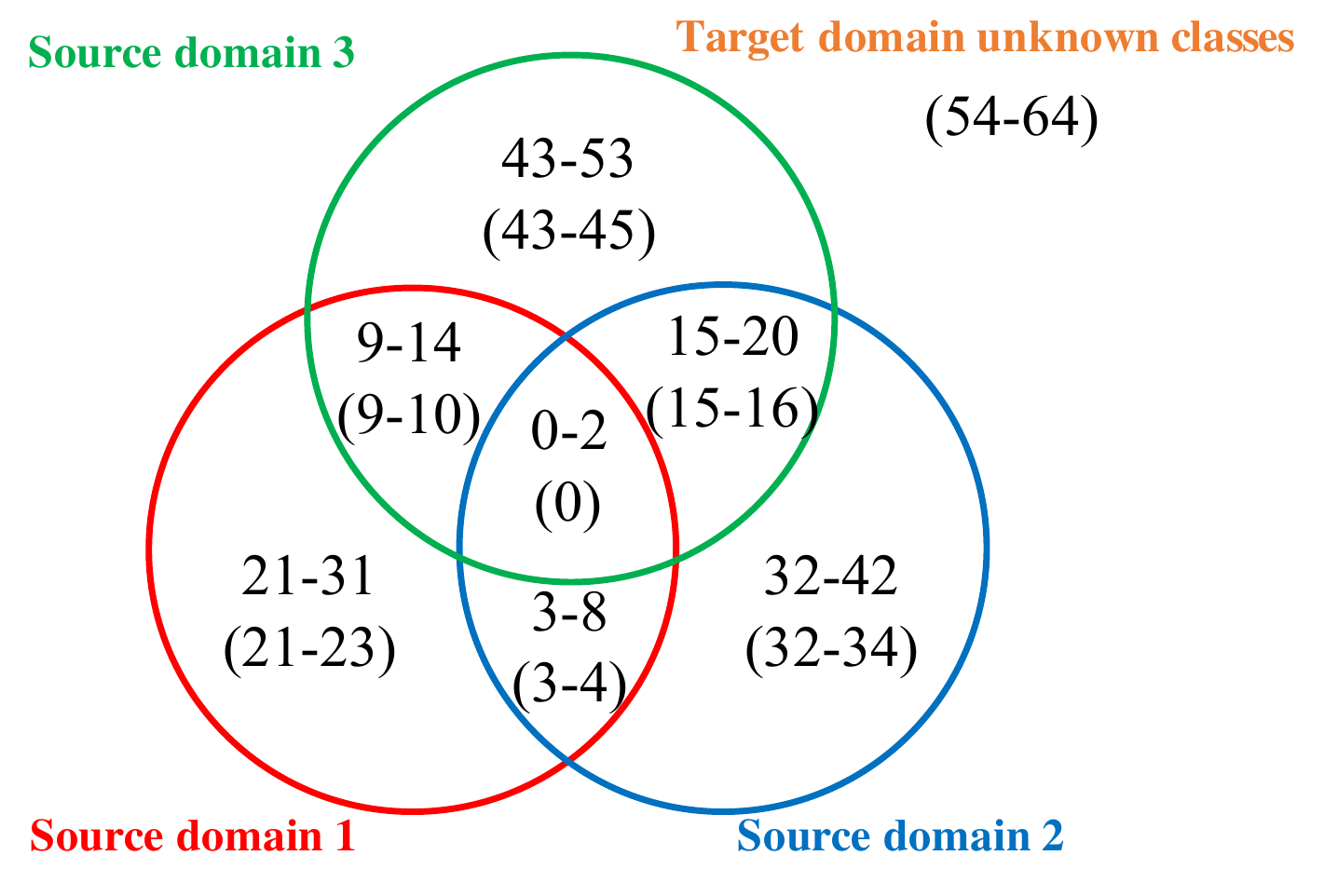}
  \vspace{-5pt}
  \caption{Illustration of the open-domain split of Office-Home Dataset. Indices without brackets show the distribution of categories among source domains, while indices in brackets indicate the categories of the target domain.}
   \label{fig:OfficeHome}
\end{figure}

\textbf{Office-Home}~\cite{cite:CVPR17OfficeHome} comprises of images from four different domains: Artistic (\textbf{Ar}), Clip art (\textbf{Cl}), Product (\textbf{Pr}) and Real-world (\textbf{Rw}). It has a large domain gap and 65 classes which is much more than other DG datasets, so it is very challenging. Similar to the PACS dataset, we spread these 65 classes among the four domains to derive an open-domain dataset and construct four open generalization tasks based on it: ArPrRw-Cl, ArClPr-Rw, ArClRw-Pr, ClPrRw-Ar, where each domain is used as the target domain respectively, and the other three domains serve as source domains. With more classes, it is possible to construct more complicated open-domain situations compared with PACS dataset. The categories contained in each domain are shown in Figure \ref{fig:OfficeHome}. 

\begin{table}[htbp]
	\begin{center}
		\caption{Open-domain split of Office-Home dataset.}
		\label{table:OfficeHomeSupp}
				\begin{tabular}{cc}
					\hline
					\textbf{Domain}&\textbf{Classes} \\
					\hline
					Source-1 &$0-2, 3-8, 9-14, 21-31$\\
					Source-2 &$0-2, 3-8, 15-20, 32-42$\\
					Source-3 &$0-2, 9-14, 15-20, 43-53$\\
					Target &$0, 3-4, 9-10, 15-16,$ \\
					&$21-23, 32-34, 43-45, 54-64$\\
					\hline
				\end{tabular}%
	\end{center}
	\vspace{-10pt}
\end{table}

\begin{table}[htbp]
	\begin{center}
		\caption{Class details in Multi-Datasets.}
		\label{table:Multi}
				\begin{tabular}{cc}
					\hline
					\textbf{Domain}&\textbf{Classes} \\
					\hline
					Office-31 &$0-30$\\
					Visda &$1, 31-41$\\
					STL-10 &$31, 33, 34, 41, 42-47$\\
					DomainNet &$0, 1, 5, 6, 10, 11, 14, 17, 20, 26$ \\
					&$31-36, 39-43, 45-46, 48-67$\\
					\hline
				\end{tabular}%
	\end{center}
	\vspace{-10pt}
\end{table}

\textbf{Multi-Datasets} scenario is constructed in this paper to consider a more realistic situation of learning generalizable representations from arbitrary source domains. We simulate the process where we obtain source domains from different resources and try to learn a generalizable model to achieve high accuracy on an unseen target domain. We leverage several public datasets including \textbf{Office-31}~\cite{cite:ECCV10Office}, \textbf{STL-10}~\cite{coates2011analysis} and \textbf{Visda2017}~\cite{visda2017} as source domains, and evaluate the generalization performance on \textbf{DomainNet}~\cite{peng2019moment}. In Office-31, we use the Amazon domain, which consists of 31 classes of office environment objects, and the images are downloaded from online merchants, which is a very popular way to acquire data. STL-10 is composed of 10 classes for general object recognition, and we use its labeled data as one of the source domains. Visda2017 dataset forms a simulation-to-real situation. We leverage its training data as the source domain, which contains synthetic images of 12 classes from CAD models. DomainNet is a new benchmark for evaluating cross-domain generalization performance. We use its four domains: Clip art, Real, Painting and Sketch as the target domains. For DomainNet, we preserve all the 23 classes existing in the joint label set of source domains and randomly sample 20 other classes as unknown classes, since there are too many open classes in it. Note that there exist huge distribution discrepancy and label-set disparity across the datasets, which forms a natural open-domain generalization scenario.

\subsection{Implementation}
We implement our algorithm in PyTorch \cite{paszke2019pytorch}. We use ResNet-18 \cite{he2016deep} pre-trained on ImageNet as the backbone network and train our model for 30 epochs with SGD as the optimization algorithm. For the proposed DAML, similar to~\cite{finn2017model}, we use fast first-order approximation to estimate gradients. To enable open-class detection for non-open-set methods, we set a confidence threshold $T$ on the prediction, where $T$ is selected similar to the open set recognition method~\cite{hassen2020learning}, by sorting the confidence on the source validation data, and then picking a certain percentile. The initial learning rate  $\beta$ is 0.001, and is decayed after 24 epochs by a factor of 10. In PACS dataset, we follow the protocol in \cite{cite:ICCV17DBA} for train and validation split. In other datasets, we randomly select 1$0\%$ data in each category of the source domains as their validation sets. We tune the hyper-parameters and choose the models for test on the held-out validation sets. We choose the step for inner update of meta-training $\eta=0.01$, and the parameters for Dirichlet mixup $\alpha_\text{max}=0.6, \alpha_\text{min}=0.2$. For DAML and all the compared methods, we use the same basic data preprocessing on the image and the same backbone. We run each experiment $3$ times and compute the average and the standard deviation.

\section{Computing Infrastructure}
We use PyTorch 1.5, torchvison 0.6 and CUDA 10 libraries. We use a machine with 32 CPUs, 256 GB memory and one NVIDIA TITAN X. The average training time for each run is 2 hours.

\section{Experiment Results}
In this section, we provide more experiment results, including the sensitivity of hyperparameters, the results of different classes, the effect of sharing parameters, and the visualization of classification results.

\begin{figure}[htbp]
    \centering
    \subfigure[$\alpha_\text{max}$]{\includegraphics[width=.23\textwidth]{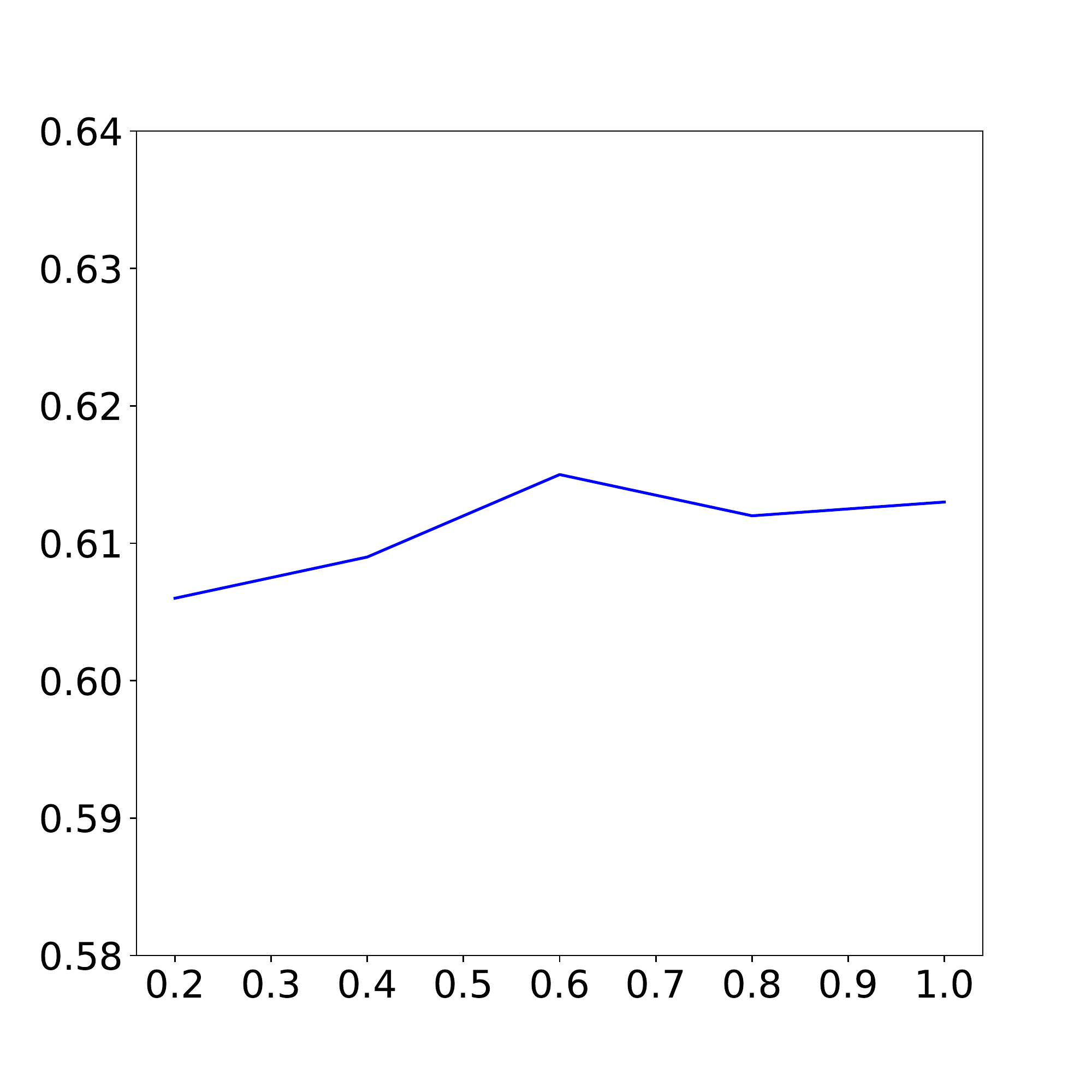}\label{fig:alpha_max}}
    \subfigure[$\alpha_\text{min}$]{\includegraphics[width=.23\textwidth]{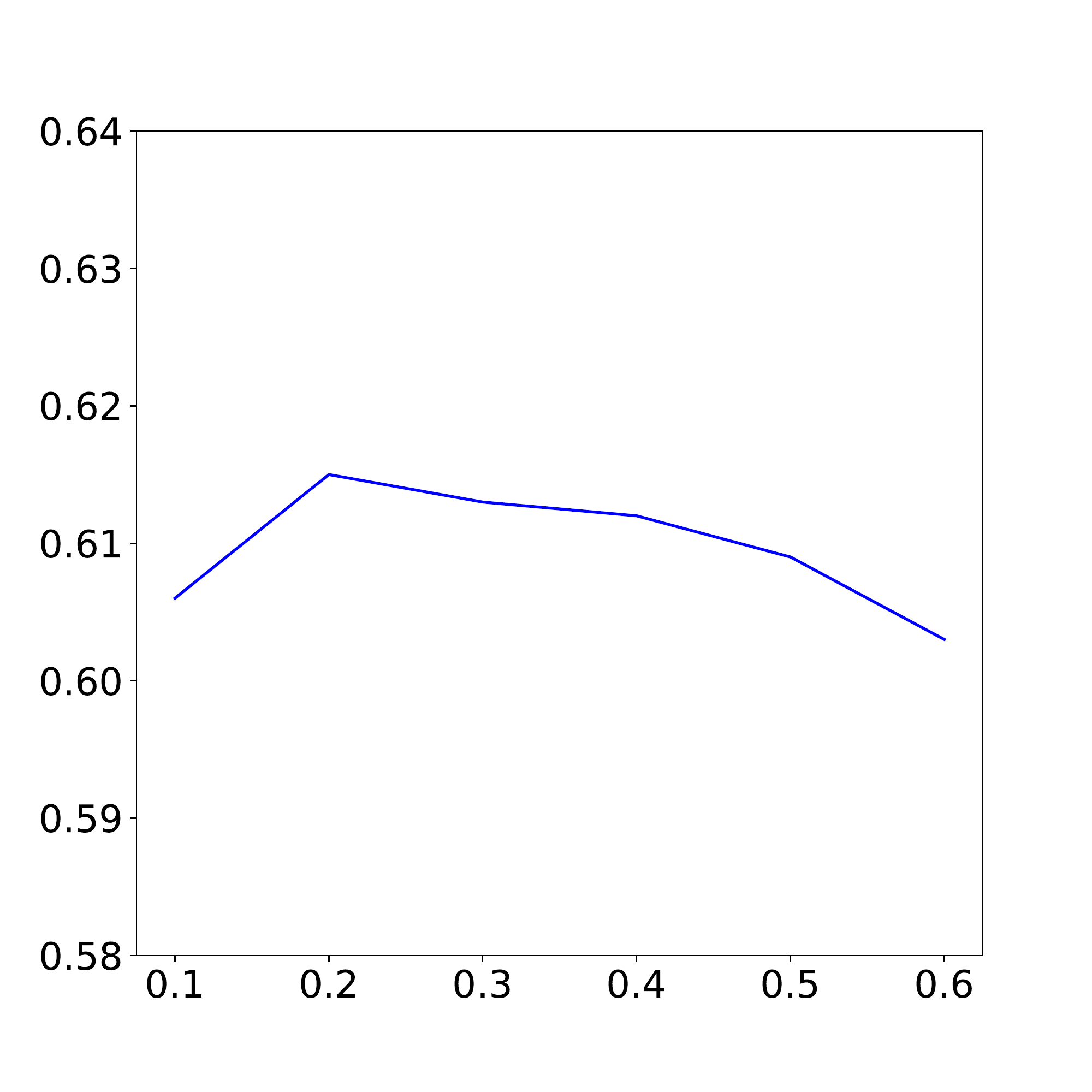}\label{fig:alpha_min}}
    \caption{Sensitivity of hyper-Parameters $\alpha_\text{max}$ and $\alpha_\text{min}$.}
    \label{fig:sensitivity_alpha}
    \vspace{-10pt}
\end{figure}

\begin{figure}[htbp]
    \centering
    \subfigure[$\beta$]{\includegraphics[width=.23\textwidth]{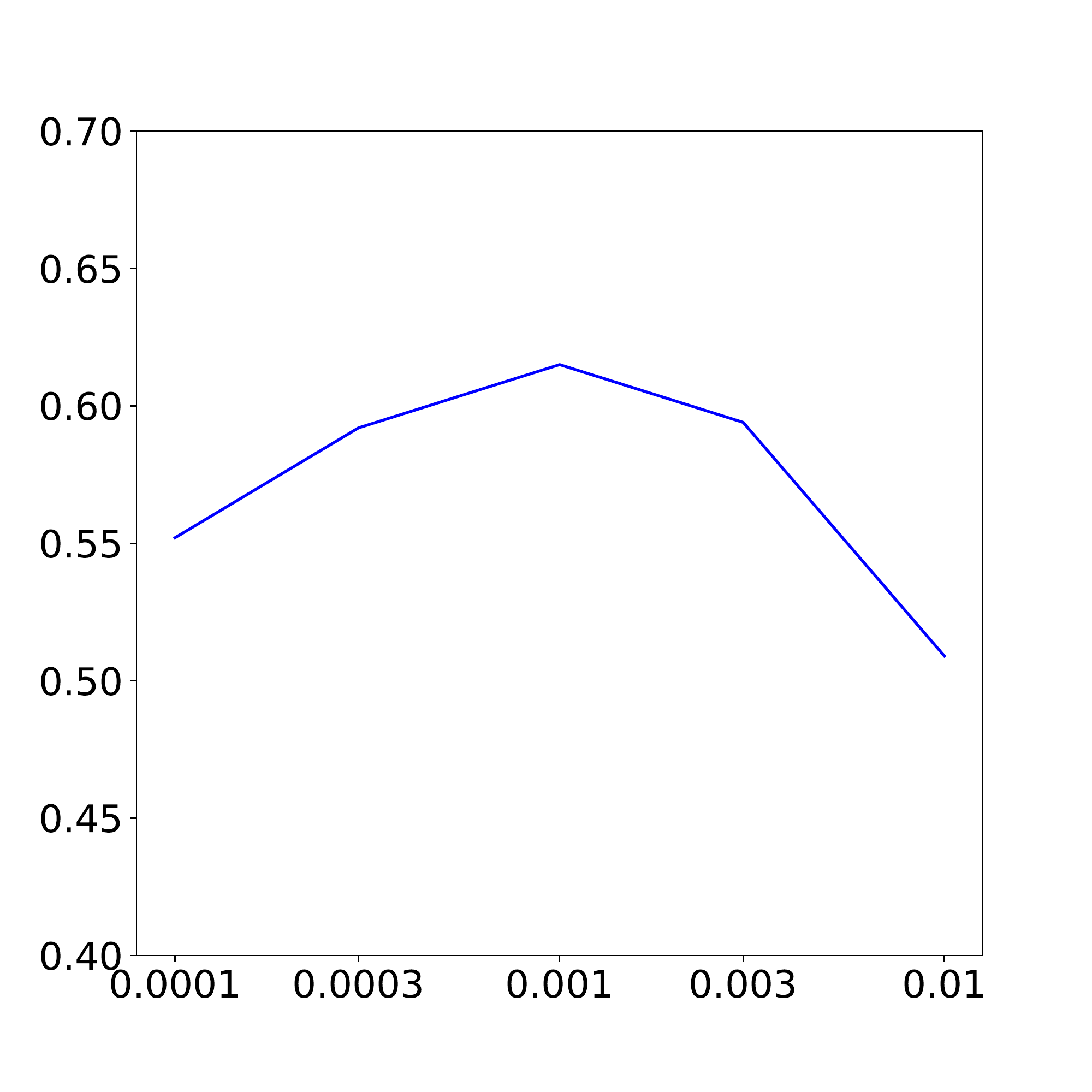}\label{fig:beta}}
    \subfigure[$\eta$]{\includegraphics[width=.23\textwidth]{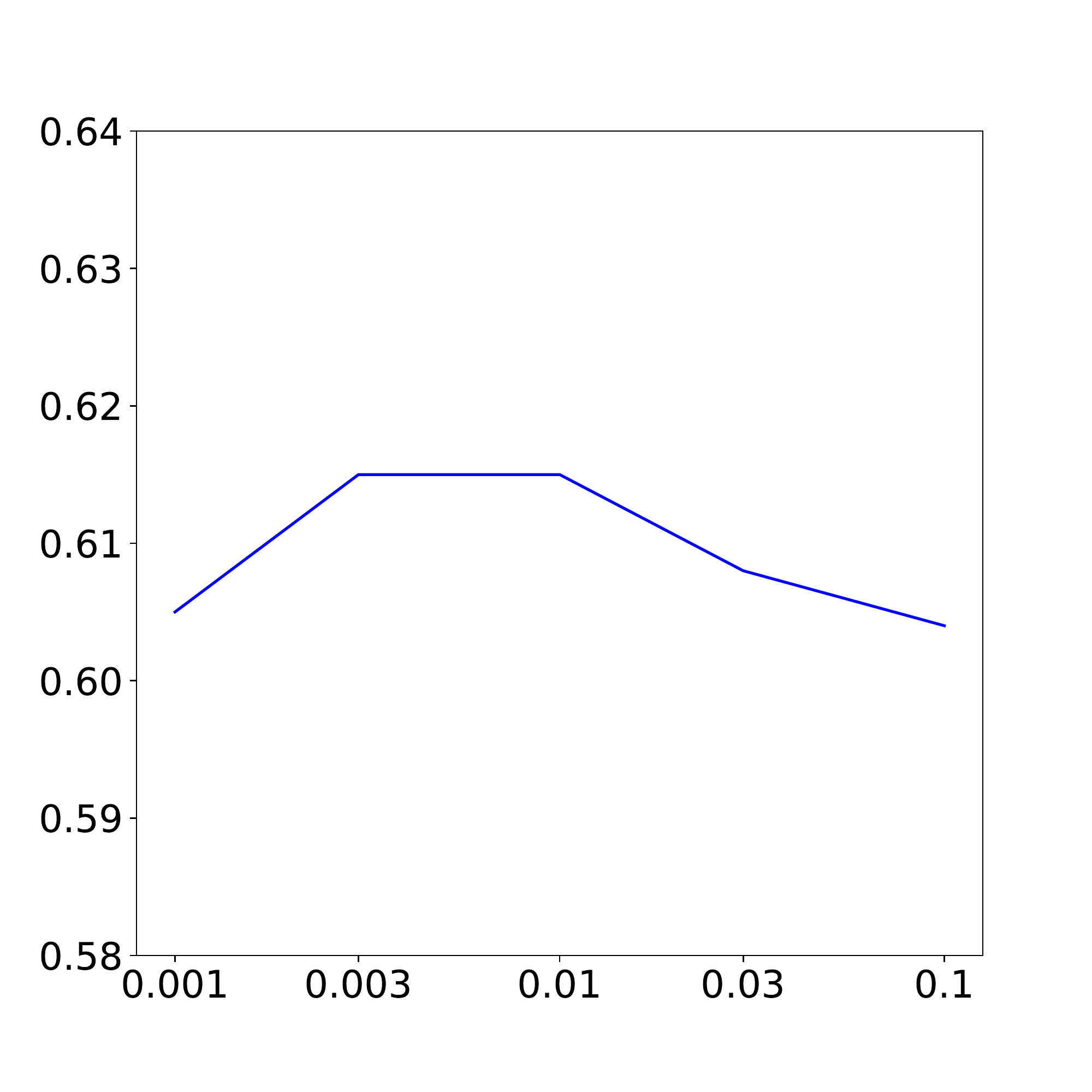}\label{fig:eta}}
    \caption{Sensitivity of hyper-Parameters $\beta$ and $\eta$.}
    \label{fig:sensitivity_beta_eta}
    \vspace{-10pt}
\end{figure}

\begin{figure*}[tbp]
    \centering
    \subfigure[Classes in 1 source domain]{\includegraphics[width=.32\textwidth]{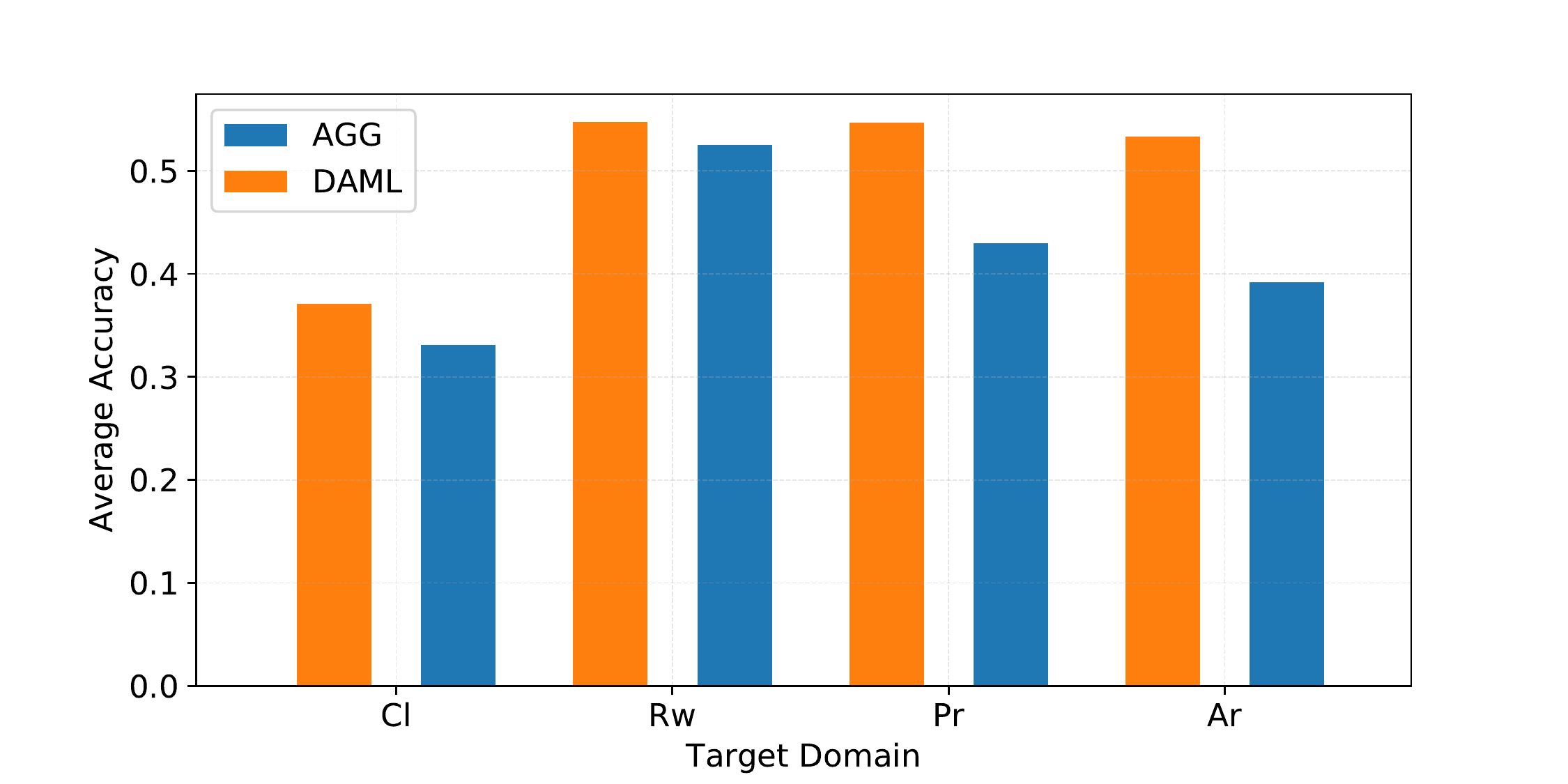}\label{fig:inlier1}}
    \subfigure[Classes in 2 source domains]{\includegraphics[width=.32\textwidth]{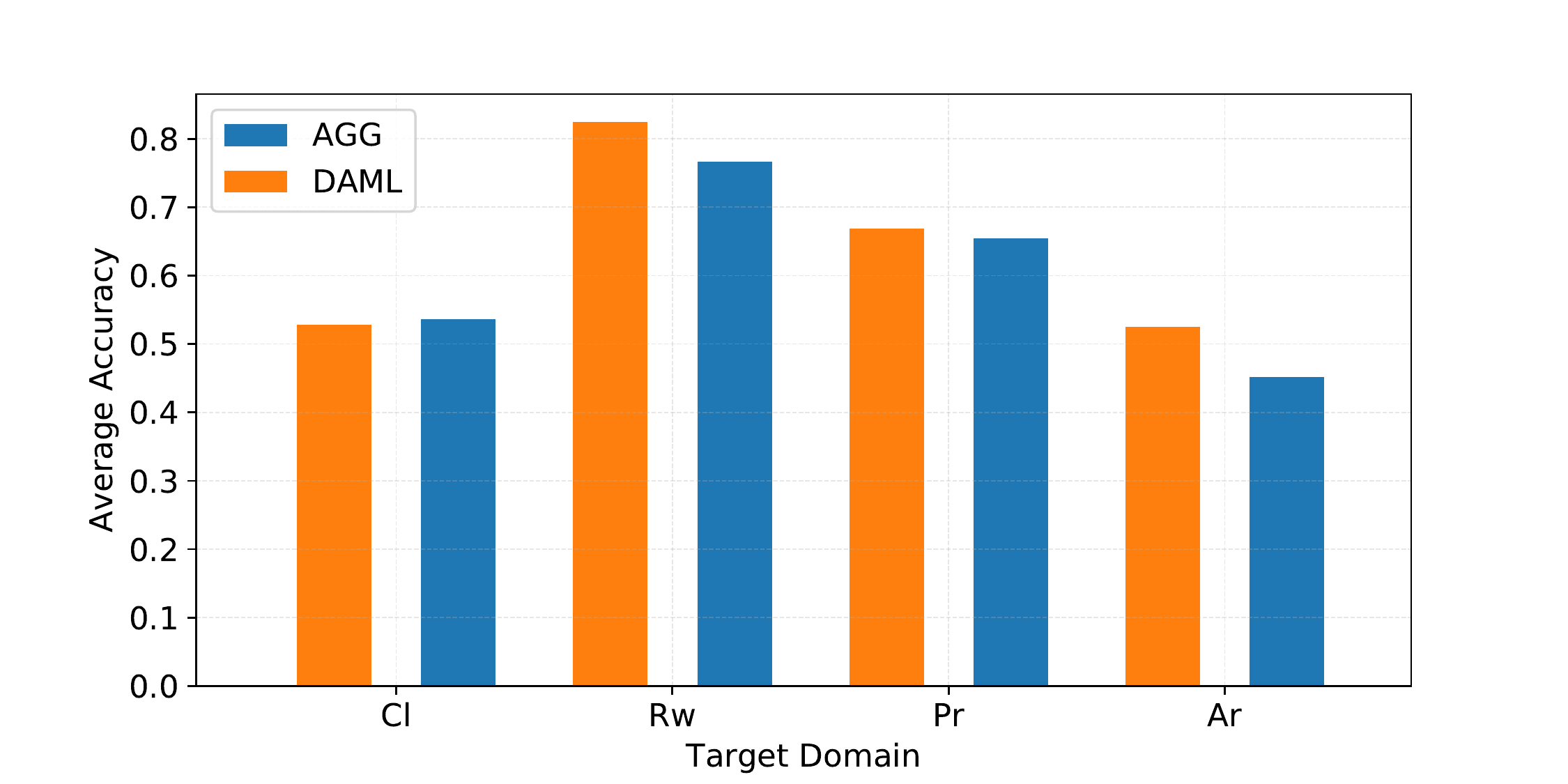}\label{fig:inlier2}}
    \subfigure[Classes in 3 source domains]{\includegraphics[width=.32\textwidth]{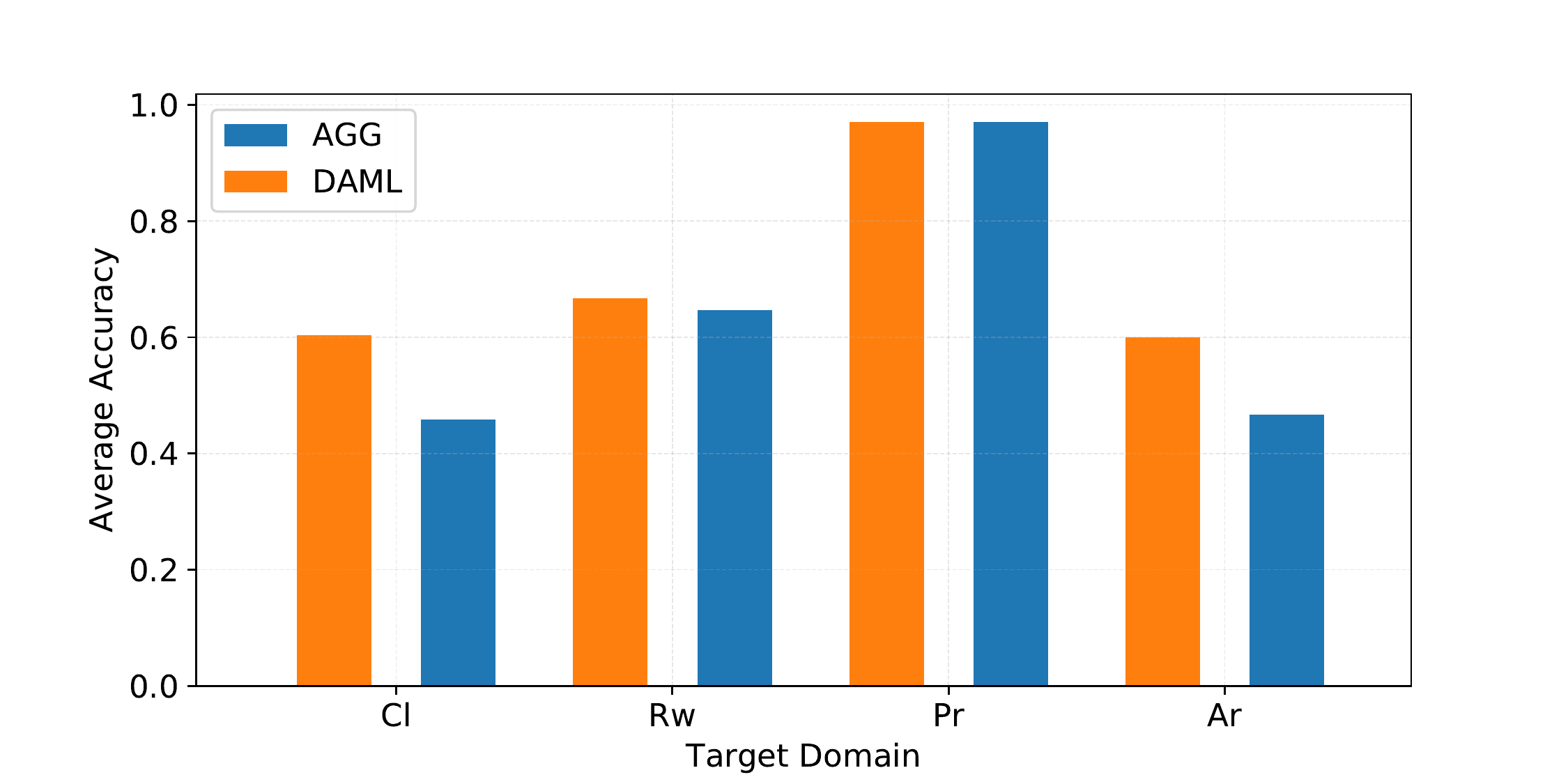}\label{fig:inlier3}}
    \caption{The average accuracy of target data from classes existing in 1 source domain, 2 source domains and 3 source domains.}
    \label{fig:inlier}
    \vspace{-10pt}
\end{figure*}

\subsection{Parameter Sensitivity}
We test the sensitivity of parameter $\alpha_\text{max}$, $\alpha_\text{min}$, $\beta$ and $\eta$. We want to demonstrate two claims: (1) The performance is stable near the optimal value of the hyper-parameters; (2) The performance will drop much when the hyper-parameters deviate from the optimal value much. The first claim demonstrates that the hyper-parameters are not sensitive and easy to tune while the second claim indicates that the hyper-parameters are still necessary even though they are not sensitive.

For $\alpha_\text{min}$, We fix $\alpha_\text{max}$ to be optimal, \textit{i.e.} $\alpha_\text{max}=0.6$ and change $\alpha_\text{min}$. For $\alpha_\text{max}$, We fix $\alpha_\text{min}$ to be optimal, \textit{i.e.} $\alpha_\text{min}=0.2$ and change $\alpha_\text{max}$. We evaluate the performance with different hyper-parameters on the DAML on ArClRw-Pr task. As shown in Figure~\ref{fig:sensitivity_alpha} and~\ref{fig:sensitivity_beta_eta}, the performance is fairly stable around the optimal value for $\alpha_\text{max}$, $\alpha_\text{min}$ and $\eta$. For $\beta$, the learning rate to finally update the parameters, the performance is stable within range $[0.0003,0.003]$, which is a widely adopted range for learning rate. On the other hand, when deviating from the optimal value a lot, the performance drops much.

\subsection{Classes with Different Domain Variations}
We have discussed in the main text that the disparate label sets between source domains cause different classes to have different domain variations. And the different domain variations lead to different performance and generalization abilities for different classes. We also argue that the previous domain generalization works fail to consider the minor class existing in few domains and thus does not perform well on such class. We empirically demonstrate the above claims in this section. 

We evaluate the accuracy of target data in four tasks of the open-domain Office-Home dataset, where each task transfers from three domains to the remaining domain. We divide the non-open target classes into three parts by how many domains each class exists in, where we have classes existing in one, two and three domains. As shown in Figure~\ref{fig:inlier}, we can observe that DAML outperforms the performance of AGG in nearly all classes, especially on the classes that exist in only one domain, which demonstrates that DAML can address the different domain variations for different classes. Also, we can observe that the accuracy of classes existing in one domain is much lower than classes in two and three domains, which demonstrates our claim on the inferior performance of minor classes.

\subsection{Trade-off between Accuracy and Efficiency}

In the ODG problem, a large domain gap exists between the source and target domains. Using a shared network for all domains is detrimental to the discriminative power on all domains. We prioritize the performance in our network design, so we use separate networks for different domains. Although using separate networks for different domains makes the training and inference time increase linearly with the number of domains, the DAML framework also allows networks of different domains to share parameters. We explore the architecture where the three domains each have a specific classifier but share the whole backbone, denoted as DAML-S. We compare DAML, DAML-S and the baseline of domain aggregation in a shared network (AGG) on the open-domain Office-Home dataset. As shown in Table~\ref{table:Shared}, the accuracy drops a little when sharing all the backbone parameters across domains, but DAML-S still outperforms the baseline with a large margin. Note that with the shared backbone, DAML-S has only a bit more per-batch training time and nearly the same per-image inference time compared with only one network. Thus, we can consider sharing parts of the network parameters across domains as a trade-off between accuracy and efficiency.

\begin{table}[htbp]
	\begin{center}
		\caption{Results on the Office-Home dataset with shared backbone.}
		\label{table:Shared}
				\begin{tabular}{cccccc}
					\hline
					\textbf{Method} &\textbf{Cl} &\textbf{Rw} &\textbf{Pr} &\textbf{Ar} &\textbf{Avg}\\
					\hline
				    AGG &$42.83$ &$62.40$ &$54.27$ &$42.22$ &$50.43$\\
					DAML&$45.13$ &$65.99$ &$61.54$ &$53.13$ &$56.45$\\
					DAML-S &$44.21$ &$64.73$ &$59.47$ &$50.81$ &$54.81$\\
					\hline
				\end{tabular}%
	\end{center}
	\vspace{-10pt}
\end{table}

\subsection{Visualization}
\begin{figure}[ht]
    \centering
    \includegraphics[width=.47\textwidth]{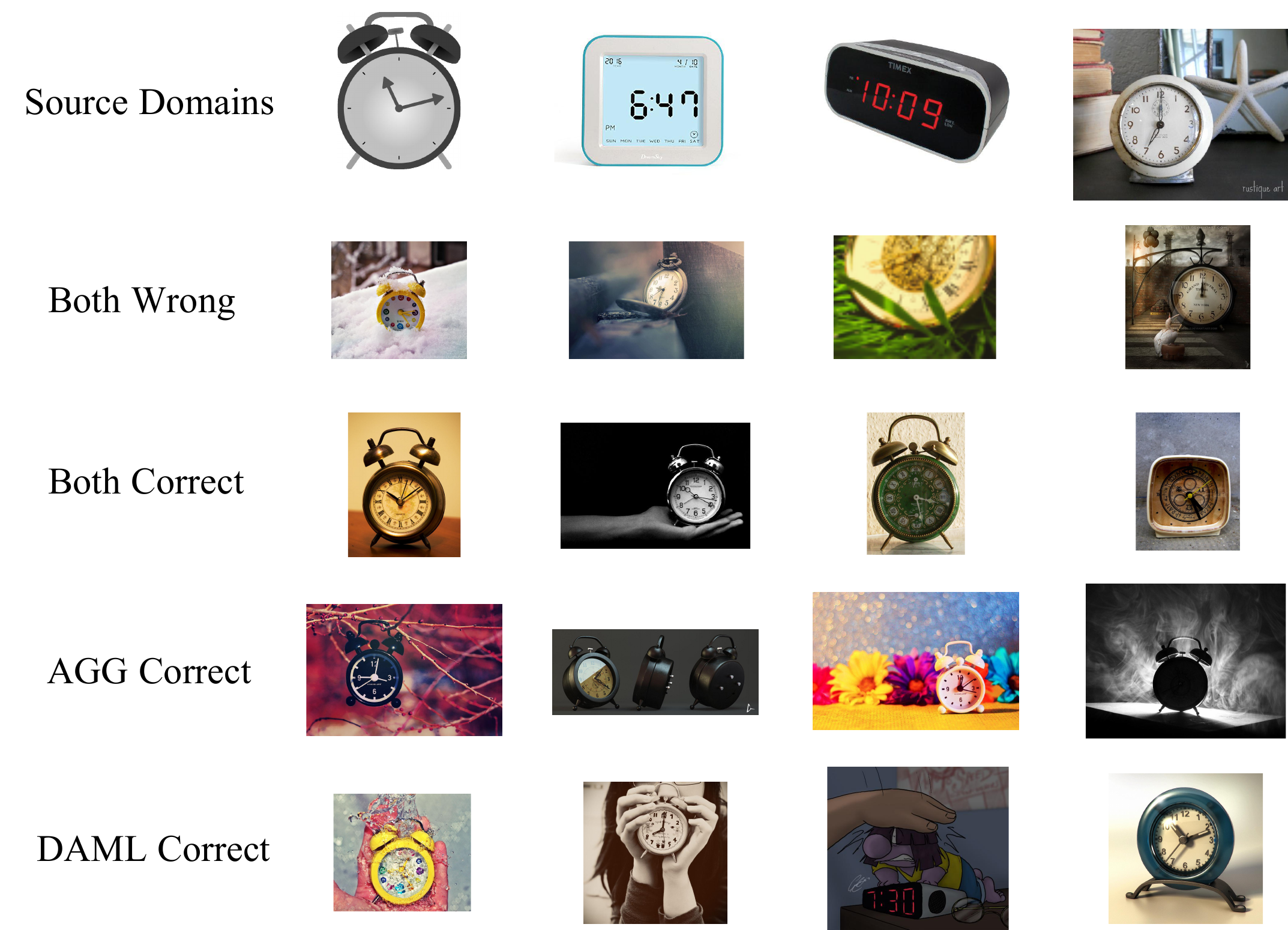}
    \caption{Visualization of classification results.}
    \label{fig:vis}
    \vspace{-10pt}
\end{figure}

We visualize the classification results of DAML and AGG on the ClPrRw-Ar task in the Office-Home dataset in Figure~\ref{fig:vis}. We visualize the source images and target images classified wrongly by both, classified correctly by both DAML and AGG, only classified correctly by AGG, and only classified correctly by DAML. We can observe that the image classified wrongly by both and only classified correctly by AGG are quite different from all the source domains, like multiple clocks and confusing background. We manually check the images only classified correctly by AGG and find that most of them are accidentally classified correctly in one run while classified wrongly in a different random seed. For the images only classified correctly by DAML, we can see a digital clock among all the mechanical clocks. The digital clock also exists in the source domains but AGG fails to learn the knowledge of them, which demonstrates that DAML can learn a more generalizable representation.

\end{document}